\documentclass[lettersize,journal]{IEEEtran}
\usepackage{amsmath,amsfonts}
\usepackage{algorithm,algpseudocode}
\usepackage{array}
\usepackage[caption=false,font=footnotesize,labelfont=sf,textfont=sf]{subfig}
\usepackage{textcomp}
\usepackage{stfloats}
\usepackage{url}
\usepackage{verbatim}
\usepackage{graphicx}
\usepackage{cite}
\usepackage{tabularx}
\usepackage{pifont}
\newcommand{\cmark}{\textcolor{Green}{\ding{51}}}%
\newcommand{\xmark}{\textcolor{Red}{\ding{55}}}%

\usepackage{fontawesome5}
\newcommand{\vehicle}{\textcolor{Blue}{\faCar}}
\newcommand{\hospital}{\textcolor{red}{\faHospital}}
\newcommand{\industry}{\textcolor{gray}{\faIndustry}}
\newcommand{\phone}{\textcolor{Black}{\faMobile}}
\newcommand{\satellite}{\textcolor{gray}{\faSatellite}}
\newcommand{\social}{\textcolor{Purple}{\faUsers}}
\newcommand{\AGI}{\textcolor{Green}{\faRobot}}
\usepackage{multirow}
\usepackage{cleveref}
\usepackage{booktabs}
\usepackage[dvipsnames]{xcolor}
\usepackage{colortbl}
\definecolor{Gray}{gray}{0.85}
\usepackage[inline]{enumitem}
\newenvironment{tableitemize}
{ \begin{minipage}[t]{\linewidth} \vspace{-10pt} \begin{itemize}[leftmargin=10pt] \vspace{5pt}}
{  \vspace{5pt} \end{itemize} \end{minipage}   }

\begin{document}

\title{Decentralized Federated Learning: \\ A Survey and Perspective}

\author{Liangqi Yuan,~\IEEEmembership{Student Member,~IEEE},
        Ziran Wang,~\IEEEmembership{Member,~IEEE},
        Lichao Sun,~\IEEEmembership{Member,~IEEE},
        Philip S. Yu,~\IEEEmembership{Life Fellow,~IEEE}, and
        Christopher G. Brinton,~\IEEEmembership{Senior Member,~IEEE}

\thanks{Manuscript received May 03, 2024.}
\thanks{L. Yuan, Z. Wang, and C. G. Brinton are with the College of Engineering, Purdue University, West Lafayette, IN 47907, USA (e-mail: liangqiy@purdue.edu; ryanwang11@hotmail.com; cgb@purdue.edu).}
\thanks{L. Sun is with the Department of Computer Science and Engineering, Lehigh University, Bethlehem, PA
18015, USA (e-mail: lis221@lehigh.edu).}
\thanks{P. S. Yu is with the Department of Computer Science, University of Illinois at Chicago, Chicago, IL 60607, USA (e-mail: psyu@uic.edu).}}



\maketitle

\begin{abstract}
Federated learning (FL) has been gaining attention for its ability to share knowledge while maintaining user data, protecting privacy, increasing learning efficiency, and reducing communication overhead. Decentralized FL (DFL) is a decentralized network architecture that eliminates the need for a central server in contrast to centralized FL (CFL). DFL enables direct communication between clients, resulting in significant savings in communication resources. In this paper, a comprehensive survey and profound perspective are provided for DFL. First, a review of the methodology, challenges, and variants of CFL is conducted, laying the background of DFL. Then, a systematic and detailed perspective on DFL is introduced, including iteration order, communication protocols, network topologies, paradigm proposals, and temporal variability. Next, based on the definition of DFL, several extended variants and categorizations are proposed with state-of-the-art (SOTA) technologies. Lastly, in addition to summarizing the current challenges in the DFL, some possible solutions and future research directions are also discussed.

\end{abstract}


\section{Introduction}
\label{Sec. Introduction}
\IEEEPARstart{F}{ederated} learning (FL) is a decentralized learning paradigm with natural privacy-preserving capabilities, which shares only model weights instead of user data \cite{kairouz2021advances}. Federated learning was first proposed by Google researchers in 2016 \cite{mcmahan2017communication} and was applied to build a language model collaboration framework on Google Keyboard to learn whether people clicked on recommended suggestions and contextual information \cite{hard2018federated}. In 2020, Google researchers expanded the concept of FL to federated analytics\cite{ramage2020federated,wang2021federated,elkordy2023federated,chen2021digital,pandey2021edge,wang2022fedfpm,froelicher2021truly}, extending from learning tasks to collaborative computing, data analysis, and inference, further deploying it within Google Keyboard. FL has demonstrated its excellent capabilities in various areas, including intelligent transportation, internet of things (IoT), healthcare, manufacturing, agriculture, energy, remote sensing, and more \cite{wang2022mobility,khan2021federated,li2024glh,li2024learning,kaissis2020secure,durrant2022role,saputra2019energy,yuan2023fedmfs,shi2022toward,pfitzner2021federated,nguyen2022federated,li2021federated,jiang2020federated,yang2022federated,chellapandi2024predictive,sun2022scalable,chu2024only,yuan2024communication}. FL also breaks geographical limitations allowing efficient collaboration worldwide \cite{tan2024bridging}. Researchers employed FL to aggregate data from 20 institutes worldwide to train a universal model to predict clinical outcomes of COVID-19 patients \cite{dayan2021federated}. FL improves the generalization capability of the model to include knowledge of diverse data. Other researchers have also used FL to aggregate data from 71 sites for rare cancer boundary detection, which greatly enriches the dataset to support research on rare diseases \cite{pati2022federated}. 

\begin{figure}[t]
\centering
\centerline{\includegraphics[width=0.8\linewidth]{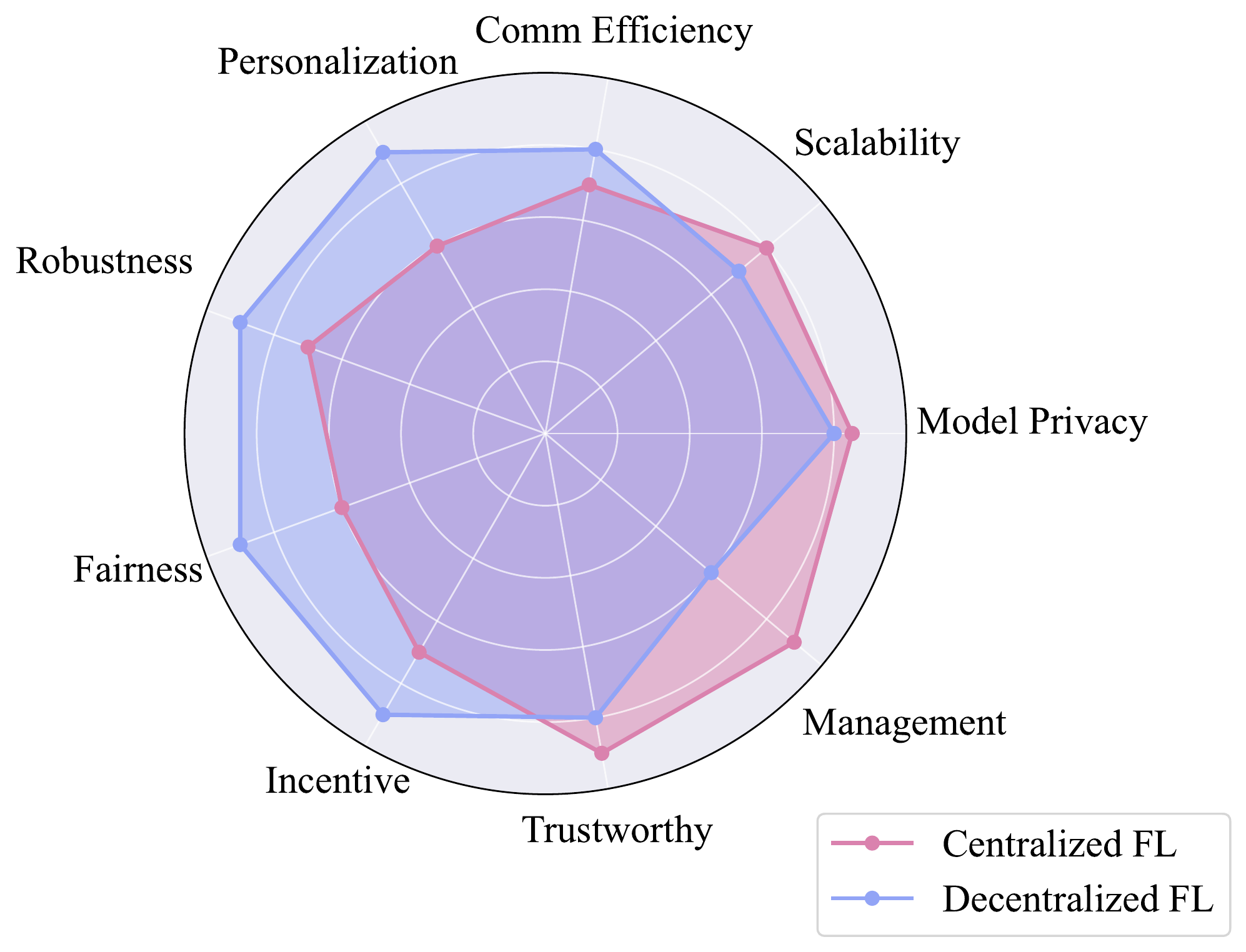}}
\caption{Comparative analysis between centralized FL and decentralized FL across various performance metrics. Each axis represents a metric with the plotted values indicating the relative strength of the respective FL approach in that domain.}
\label{Fig. Comparison}
\end{figure}

Traditional FL focuses on the decentralized learning and centralized aggregation paradigm established by data parallelism. Data parallelism refers to the situation where the raw data of the clients is generated in parallel locally, and this raw data is neither sent out nor visible to others. Each client trains a model based on its local data and then communicates the model parameters with the server to ensure the effective integration of learning results from each client and obtain a global model. An FL taxonomy refers to the number and nature of clients participating in the learning network, including cross-silo and cross-device FL frameworks \cite{kairouz2021advances}. The clients in cross-silo FL usually are different organizations, research institutions, data centers, etc., which may have more reliable communication, computational resources, and a large amount of data. The clients in cross-device FL are huge mobile or IoT devices, which can encounter potential bottlenecks in communication and computation. Another FL taxonomy is considered for differences in data distribution among clients, including horizontal, vertical, and transfer \cite{yang2019federated}. In horizontal FL, clients have more similar sample features and fewer identical users. Clients in vertical FL have more similar users and fewer similar sample features. Federated transfer learning clients have neither many similar sample features nor similar users.

\begin{figure*}[t]
\centering
\centerline{\includegraphics[width=\linewidth]{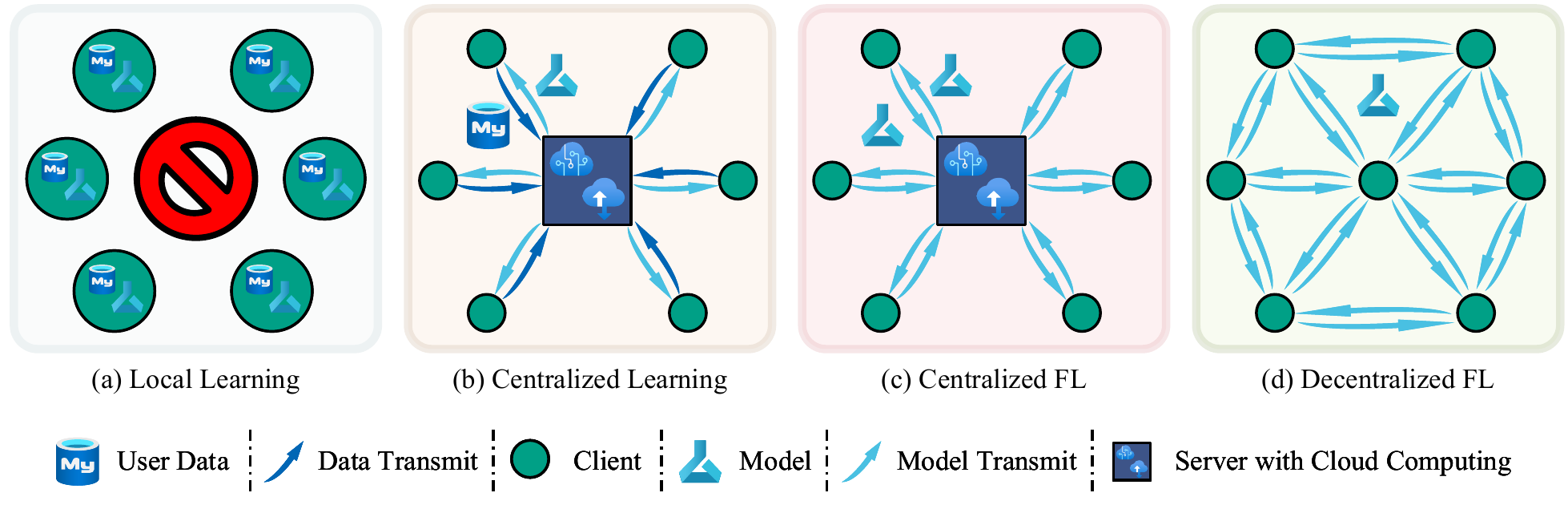}}
\caption{Illustration of local learning, centralized learning, CFL, and DFL. (a) Clients are trained with local user data only. The clients neither share raw data nor communicate with each other. (b) After clients send the user data packets to the server, the server trains a general model using all the data. The generalized model is then shared with all clients. (c) Clients send the locally trained model parameters to the server. The server aggregates all the local models and then transmits the aggregated global model parameters to all the clients. (d) Clients share their locally trained model with other clients. Subsequent clients then continue to learn, personalize, and adapt the model locally, while also exchanging and propagating the model parameters that possess local knowledge.}
\label{Fig. FL}
\end{figure*}

In this paper, we present a thorough investigation into decentralized FL (DFL) and offer novel perspectives on its taxonomies. Distinguishing from the conventional centralized FL (CFL) that relies on a central server for aggregation, we specifically focus on the less-explored DFL framework, which operates independently of a central server. Fig. \ref{Fig. Comparison} illustrates a comparison between CFL and DFL across nine key evaluation metrics, which are the focal points of current state-of-the-art (SOTA) research and worthy of further investigation \cite{yu2020fairness,chen2021communication,yang2022trustworthy,zhang2021faithful,bonawitz2019towards,li2021survey,lai2022fedscale,agrawal2022federated,antunes2022federated,kang2020reliable,lyu2020threats,zhan2020learning,blanco2021achieving,yu2020sustainable,wan2021convergence,ghosh2019robust}. Given their inherent characteristics, CFL and DFL each exhibit unique advantages in different applications.

Fig. \ref{Fig. FL} shows the illustration of local learning, centralized learning, CFL, and DFL. In the local learning strategy, the user data and trained model of each client are only used locally, and they do not communicate with any other clients or servers, as shown in Fig. \ref{Fig. FL}(a), but this may lead to overfitting. Alternatively shown in Fig. \ref{Fig. FL}(b), the centralized learning strategy involves the transmission of raw data in the communication between clients and the server, which consolidates and centralizes the learning process but does not guarantee the privacy of the users. Both of these strategies are often used by researchers as baselines to compare with FL.

CFL is a centralized structure where a server will communicate, coordinate, and manage all clients. Fig. \ref{Fig. FL}(c) shows the communication between clients and the server. Clients learn on local data and then upload the trained model parameters to the server. The server aggregates the local models and then shares the global model with the clients. The idea is that all clients contribute to one global model, and the one global model is applied to all clients. For CFL, clients only share the trained local model parameters with the server but not the users' raw data. FL not only protects users' privacy and improves learning efficiency, but also saves communication resources when the model size is much smaller than the data size. 

DFL is a decentralized structure in which clients communicate and share model parameters with each other without any server. There are relevant designations in the recent literature, such as peer-to-peer FL \cite{lalitha2019peer}, server-free FL \cite{he2019central}, serverless FL \cite{he2021spreadgnn}, device-to-device FL \cite{xing2021federated}, swarm learning \cite{warnat2021swarm}, etc. Fig. \ref{Fig. FL}(d) shows clients communicating directly with other clients without server coordination. Since there is no unified coordination and configuration of servers, the communication network between clients is more diverse. For the DFL discarding the server is considered to be more customizable, which can further save communication and computational resources with higher confidence in diverse variants. The pointing and peer connections in the communication network are adaptively configured and changed according to the scenario, which is one of the advantages of DFL. In addition to the typical line, ring, and fully connected peer connection types, it is conceivable to connect based on geographical neighbors, the similarity of clients, communication protocols, etc. 

The concept of DFL was first proposed in the year 2018 \cite{lalitha2018fully}. As of June 1, 2023, a search on Google Scholar yields 1,350 results related to DFL, with a substantial number of 652 contributions coming from the year 2022 alone. The research associated with DFL exhibits a persistent exponential growth trajectory. DFL has received extensive attention as an emerging framework \cite{li2020practical,savazzi2020federated,monschein2021towards,savazzi2021opportunities,shi2023improving,chellapandi2023convergence}. The most significant advantage of DFL is that it eliminates the communication resources needed for the server as an intermediary step and the high bandwidth requirements associated with it. Xu \textit{et al.} \cite{xu2021federated} listed DFL, model compression, selective client communication, and low communication frequency as four ways to reduce communication costs. Lian \textit{et al.} \cite{lian2017can} demonstrated the advantages of decentralized learning over centralized learning, especially since the number of clients in decentralized learning is proportional to the speedup. 

\begin{table*}[t]
\renewcommand{\arraystretch}{1.4}
\small
\caption{Comparison of Related Surveys of Decentralized Federated Learning}
\label{Table Comparison of related survey}
\begin{tabular}{|l|l|l|l|l|l|l|l|l|}
\hline
\textbf{Survey} & \textbf{Time} & \textbf{Focused DFL Topic} & \textbf{Application Scenarios} & \textbf{Order} & \textbf{Protocol} & \textbf{Topology} & \textbf{Paradigm} & \textbf{Variability} \\
\hline
\hline
\cite{kairouz2021advances} & 2021 & Introduction & \xmark & \xmark & \xmark & \xmark & \xmark & \xmark \\
\hline
\cite{yin2021comprehensive} & 2021 & Introduction & \xmark & \xmark & \xmark & \xmark & \xmark & \xmark \\
\hline
\cite{qu2021decentralized} & 2021 & UAV & \satellite & \xmark & \xmark & \xmark & \xmark & \xmark \\
\hline
\cite{nguyen2021federatedSurv} & 2021 & IoT & \vehicle { } \industry { } \hospital { } \satellite { } City & \xmark & \xmark & \xmark & \xmark & \xmark \\
\hline
\cite{qu2022blockchain} & 2022 & Blockchain & \xmark & \xmark & \xmark & \xmark & \xmark & \xmark \\
\hline
\cite{gupta2022survey} & 2022 & Survey & \xmark & \xmark & \xmark & \xmark & \xmark & \xmark \\
\hline
\cite{witt2022decentral} & 2022 & Survey \& Reflection & \vehicle { } IoT {} Finance & \xmark & \xmark & \xmark & \xmark & \xmark \\
\hline
\cite{gao2023decentralized} & 2023 & Tutorial & \xmark & \xmark & \xmark & \xmark & \xmark & \xmark \\
\hline
\cite{gabrielli2023survey} & 2023 & Survey & \xmark & \xmark & \xmark & \xmark & \xmark & \xmark \\
\hline
\cite{wu2023topology} & 2023 & Topology & \xmark & \xmark & \xmark & \cmark & \xmark & \cmark \\
\hline
\cite{chellapandi2023federated} & 2023 & CAV & \vehicle & \xmark & \xmark & \cmark & \xmark & \xmark \\
\hline
\cite{beltran2023decentralized} & 2023 & Survey \& Tutorial & \vehicle { } \industry { } \hospital { } \phone { } \satellite { } & \xmark & \cmark & \cmark & \xmark & \xmark \\
\hline
\cite{hallaji2024decentralized} & 2024 & Security and Privacy & \xmark & \xmark & \xmark & \xmark & \xmark & \xmark \\
\hline
\hline
\textbf{Ours} & 2024 & Survey \& Perspective & \vehicle { } \industry { } \hospital { } \phone { } \satellite { } \social { } \AGI & \cmark & \cmark & \cmark & \cmark & \cmark \\
\hline
\end{tabular}
\\[2pt] 
\cmark {} Included, \xmark {} Not mentioned.
\\[2pt] 
\vehicle { }: Connected and Automated Vehicles (CAVs), 
\hospital { }: Healthcare, 
\industry { }: Industrial IoT (IIoT), 
\phone { }: Mobile Services, 

\satellite { }: Unmanned Aerial Vehicle (UAVs) and Satellites, 
\social { }: Social Networks,
\AGI { }: Artificial General Intelligence (AGI)
\end{table*}

\begin{figure*}[t]
\vspace{5pt}
\centering
\centerline{\includegraphics[width=0.85\linewidth]{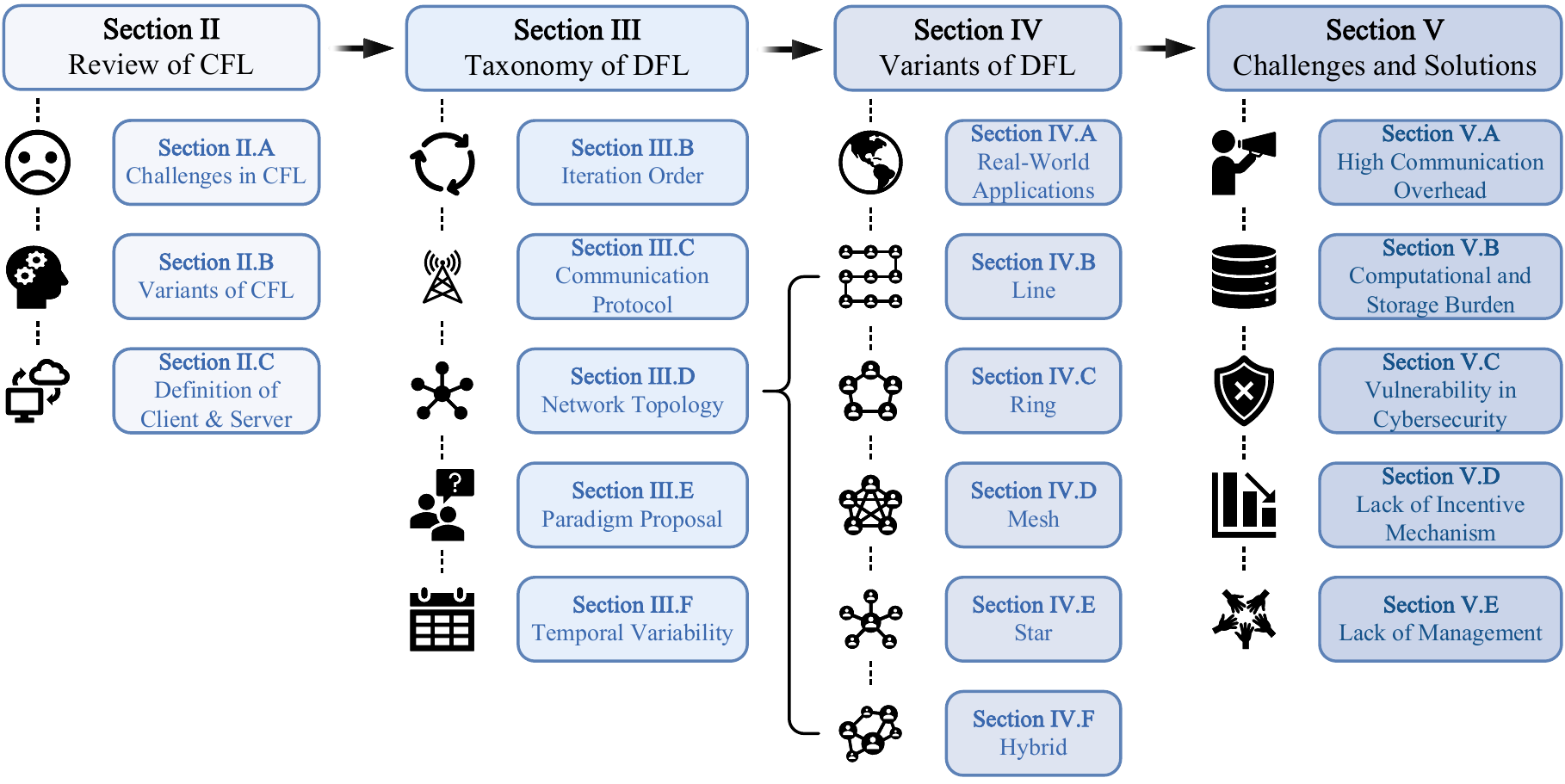}}
\caption{Roadmap for this perspective paper.}
\label{Fig. Roadmap}
\vspace{-5pt}
\end{figure*}

Although FL has shown unprecedented advantages, most of the current research has been limited to CFL. DFL, as an essential branch in FL, is proliferating and offering benefits over CFL. Recent surveys have focused more on CFL, with less attention given to DFL \cite{rieke2020future,xu2021federated,mothukuri2021survey,chellapandi2023survey}. Furthermore, there is a lack of a comprehensive, in-depth, and insightful survey that establishes the logic of building a DFL system, including iteration, communication protocol, network topology, paradigm, and more. This paper begins with a review of CFL, summarizing its challenges and various extended variants as potential solutions that can be compared and analogized with DFL. As an emerging field, this survey aims to fill gaps in the DFL survey literature by covering perspective papers that are currently not included. It systematically integrates and categorizes the SOTA research in DFL. A detailed and comprehensive comparison of our survey with other related DFL surveys can be found in Table~\ref{Table Comparison of related survey}. 

The contributions of this paper are:
\begin{itemize}
\item We provide a description of CFL, summarize the challenges, and offer a detailed introduction to the various variants, their roles, addressed issues, and advantages.
\item We systematically define and describe five taxonomies of DFL, including iteration order, communication protocol, network topology, paradigm proposal, and temporal variability. To the best of our knowledge, this is the first comprehensive and insightful perspective paper for DFL.
\item Based on the network topology, we propose and envision five variants of DFL to categorize the recent literature, anticipate potential application scenarios, and highlight the advantages.
\item We summarize five current challenges, possible solutions, and future research directions for DFL.
\end{itemize}

The presentation of this paper is summarized as shown in Fig. \ref{Fig. Roadmap}. Section \ref{Sec. Centralized FL} reviews the history of CFL, the existing challenges, and some variants as potential solutions. Section \ref{Sec. Decentralized FL} provides the definitions and descriptions of DFL communication protocol, network topology, and paradigm proposal. Section \ref{Sec. Variants of DFL} demonstrates several variants in DFL, followed by Section \ref{Sec. Challenge in DFL} analyzing the challenges of DFL. 
Finally, Section \ref{Sec. Conclusion} provides a summary of this paper.

\section{Review of Centralized Federated Learning}
\label{Sec. Centralized FL}
McMahan \textit{et al.} \cite{mcmahan2017communication} proposed the first mature and most popular FL algorithm, federated averaging (FedAvg). At each communication round, clients upload their trained local models to the server, and the server weighted averages all local models according to the number of client samples. Based on FedAvg, various derivation and optimization schemes exist to address the challenges in the FL algorithm \cite{karimireddy2020scaffold,zhao2018federated}. Li \textit{et al.} \cite{li2020federatedfedprox} developed an advanced algorithm FedProx to penalize the bias of the local model to the global model by a proximal term. The advantage is to limit the significant variance and unstable convergence of local models due to overfitting on clients with system heterogeneity. Wei \textit{et al.} \cite{wei2020federated} took into account the privacy leakage concern of model parameters uploaded by clients in FL and proposed to improve the differential privacy by adding noise before the client sends it to the server for aggregation. Also, the game trade-off between FL convergence and privacy preservation and the optimal communication rounds were highlighted.

Although the diverse derivations that exist complement the performance of FL, there are undeniable drawbacks, such as a single point of failure (SPoF) on the server. In this section, after presenting some of the challenges and limitations of the server, we show some variants of the solution and SOTA technologies.

\subsection{Challenges in Centralized Federated Learning} 
\label{Sec. Challenges in CFL}
For CFL, the server takes on many responsibilities and challenges, with large service providers, such as large organizations and research institutions, playing the role of server. While these large providers have unparalleled resources compared to small workshops, there are some concerns here as the number of clients grows endlessly \cite{gadekallu2021federated}.

\textit{1) \textbf{Client Heterogeneity}} predominantly stems from three latent factors: individual, group, and systemic heterogeneity \cite{li2020federatedfedprox,ghosh2020efficient,long2023multi,ye2023heterogeneous}. Individual heterogeneity arises from differences inherent to each sample or individual, such as variances among patients in a hospital setting. Group heterogeneity is rooted in shared characteristics among subsets of samples, such as patients of different age groups, regions, or medical backgrounds. Systemic heterogeneity originates from variations introduced during data collection by the system, which can include discrepancies from different equipment, clinical practices, or data collection personnel. 

\textit{2) \textbf{Communication Resource}} is limited on both the server and client sides \cite{fang2021over,fang2022communication}. Although FL has dramatically reduced the consumption of communication resources by sharing only model parameters instead of user raw data, communication resources are a serious problem considering the large number of parallel clients (up to one billion). In particular, when delays in communication cause the server to wait for clients with communication problems, it can also cause the whole FL framework to become highly inefficient. Some FL communication proposals have been proposed to improve communication efficiency \cite{konevcny2016federated,sattler2019robust,lan2023communication,lan2023improved}.

\textit{3) \textbf{Computational and Storage Resource}} on the server side are also challenged \cite{imteaj2021survey,wang2019adaptive,imteaj2022federated}. The server needs to store and aggregate the models of these billions of clients. Even though lightweight models are emerging recently \cite{sandler2018mobilenetv2,tan2019efficientnet}, the need to compute and store model data can easily reach petabytes in size \cite{sattler2019sparse}. Besides the current version of the massive local model, subconditionals and versioned storage of the global model may also be required. Additionally, as clients demand real-time processing of a large volume of inference tasks, this places high demands on the computational resources required during inference \cite{zhang2021deep}.

\textit{4) \textbf{Fairness, Security, and Trustworthy}} have always been crucial concerns in CFL, with these factors significantly impacting the system's overall reliability, user confidence, and data integrity. A series of questions related to security and trust form the chain of suspicion: whether the server aggregation model is reasonable, whether the global model will have high performance across all clients, whether the global model is validated, how to use the global model securely, and whether the server is secure from attacks \cite{ma2020safeguarding}. For security issues, there are different directions of research, including malicious attacks \cite{bagdasaryan2020backdoor}, data poisoning \cite{tolpegin2020data}, anomaly detection \cite{mothukuri2021federated}, and privacy protection \cite{wei2020federated}. For trustworthy \cite{zhang2023survey}, fairness \cite{li2019fair}, incentive \cite{kang2019incentive}, and interpretability \cite{roschewitz2021ifedavg} in FL are also worthy research directions.

\textit{5) \textbf{Unreliable Connection}} in FL can stem from factors such as unreliable communication conditions, malicious attacks, or server malfunctions, leading to delays, packet loss, or noise in model transmission \cite{salehi2021federated,ganguli2023fault,ganguli2023internet}. As all clients typically communicate with a central server, an SPoF can halt the entire system's update process. Although employing multiple edge servers can distribute the risk of SPoF, it may still cause the system segments connected to an affected edge server to become unresponsive. \cite{kim2019blockchained,qu2020decentralized} have explored using blockchain technology to mitigate SPoF by replacing the central server role, but this approach diverges from the conventional centralized FL model.

\subsection{Variants of Centralized Federated Learning}
\label{Sec. Variant of CFL}
The network variants and extensions of CFL are designed to address the above challenges and adapt to different real-world application scenarios.

\textit{1) \textbf{Hierarchical FL}} features a classic and popular client-edge-cloud architecture, where model parameters are infrequently transmitted between the edge and the cloud, effectively reducing communication overhead \cite{zhang2021spectrum,zhang2021optimizing,yang2023detfed}. It usually perform additional aggregations by setting up additional edge servers \cite{lim2020federated}, which aim to spread the communication \cite{ye2020edgefed} and computing pressure and reduce the impact of SPoF. These additional edge servers are geographically closer to clients, resulting in less communication resource consumption and lower latency \cite{wang2023towards,fang2023submodel,chen2023taming}. After one or more edge server aggregations, the edge servers then upload the edge global model to the cloud for aggregation into a global model. In addition to communication optimizations, the geographic proximity of edge servers to clients may also lead to better adaptation of edge servers to the connected clients. Edge servers and connected clients can be considered geographically personalized clusters. For example, by assigning edge servers to states in the United States, the state edge servers can be more personalized to the state's user scenarios and user habits, such as weather, number of users, time zone, ethnicity, age distribution, etc. 


\textit{2) \textbf{Personalized FL}} can be classified into two categories, i.e., global model personalization and personalized model architecture \cite{chu2022mitigating,tan2022towards,chu2022multi}. Global model personalization usually starts with a global model, and then the client personalizes this global model to fit the local user. Personalization is the behavior of the client independent of the server, such as federated transfer learning to transfer global model knowledge locally \cite{chen2020fedhealth,yuan2023federated}. The personalized model architecture changes the traditional FL architecture to develop a personalized model with user knowledge, which is the behavior of the server. A famous architecture is clustered FL that has been of interest to researchers \cite{sattler2020clustered,ghosh2020efficient}. The client model in the personalized FL framework is closer to the user, so it is known for its high accuracy and confidence. In particular, it is a highly effective solution for non-independent and identically distributed (non-IID) data. When the aggregated global model deviates from the user, personalization can transfer the model and adapt it to different heterogeneities.

\textit{3) \textbf{Split FL}} splits the model for learning, where the server is responsible for some model layers \cite{thapa2022splitfed,han2023splitgp,han2023federated}. The only data sent by the client to the server are the hidden representations and/or gradients in the cut layer of the model. The client not only shifts part of the learning task to the server but also does not share the user data. Compared to traditional FL, the split FL framework has similar accuracy and communication efficiency with a lower learning burden on the client side and more robust privacy protection. However, split FL is still in its early stages and has significant limitations, such as the need to consume more communication resources. Especially the presence of SPoF on the server can have even greater consequences.

\textit{4) \textbf{Graph FL}} is particularly effective for applications involving graph-structured data, such as social networks, transportation networks, molecular structures, etc \cite{qu2023semi,he2021fedgraphnn}. This effectiveness stems from the fact that clients possess local or independent graph data, which includes rich relational information about nodes and edges. More specifically, Graph FL encompasses three levels, depending on the level of detail clients have about the graph data. These levels include multiple graphs, different parts of multiple graphs, and parts of a single graph \cite{baek2023personalized}. In addition to employing graph neural networks for graph data in FL, some researchers have also considered topological graphs between clients \cite{liu2024federated}. These topological graphs can be established based on various factors, exploring network connectivity conditions between clients and the server, data and model availability of clients, as well as similarities and data generality among clients \cite{wagle2022embedding,wagle2023reinforcement,wagle2024unsupervised}.

\textit{5) \textbf{Asynchronous FL}} is designed to overcome the limitations of FL frameworks that require clients to synchronize model updates. This approach aligns well with the real-world scenarios in which the client's data updates and model training vary significantly. In these systems, heterogeneous clients have different amounts of data and computational resources and can train and update their local models at any time without the server having to wait for all clients to synchronize \cite{lan2024asynchronous,chang2024asynchronous}. Asynchronous FL is particularly suitable for environments with dynamic changes, a large number of clients, and a wide distribution, such as mobile devices, which may conduct model training during idle times rather than during routine user interactions \cite{bian2024accelerating,zhou2024adaptive}.

Variants of CFL currently exist with exotic frameworks that may include single, multiple, sub, and master servers to optimize and target different problems. 
For example, Zhang \textit{et al.} \cite{zhang20242} proposed the (Com)$^2$Net, a large-scale distributed computing framework capable of spanning space-air-ground, end-edge-cloud, and multi-data center environments, facilitating ubiquitous connectivity and collaborative computing in FL. In addition to various variants, a popular approach is to assemble various variants of the FL framework to target multiple issues \cite{chen2021privacy}.

\subsection{Definition of Client and Server}
\label{Sec. Definition of Client and Server}

This paper proposes a new taxonomy that focuses on the roles played by communication endpoints in FL. We argue that the roles of edge devices, compute clusters, institutions, and organizations are relative rather than absolute. For example, a university may act as a central server when managing edge devices within its campus, but it should be considered as a client when communicating with other universities. The classification of a role as a client can be determined by whether it generates raw data and stores the data locally. For instance, a healthcare institution both generates clinical data for patients and retains it in its own database without sharing it with other institutions. Additionally, clients and servers are not mutually exclusive roles. A healthcare institution, for example, can act as both a client and a server \cite{pati2022federated}. It generates raw patient data and performs local training while also serving as a server by receiving model parameters from other healthcare institutions, aggregating them, and sharing the updated model. In practice, it can be considered as a star topology network in DFL.

The emphasis on a star topology network instead of CFL is driven by the more pressing issues of fairness and trust in FL systems when an institution simultaneously possesses raw data for local training and assumes the role of a server for aggregation. Ensuring that this institution does not favor its local data during aggregation poses an open question. As the saying goes, One cannot be a judge, jury, and executioner. Similarly, an institution should not act as both a client and a server. However, this situation is more common in the real world. On one hand, institutions with more raw data have greater influence and often initiate tasks. On the other hand, institutions with more data also tend to have more abundant computing resources, making them better suited for the server role. In today's world, where large institutions possess more resources and hold greater influence, addressing the interests, privacy, and fairness of non-representative clients is a challenge.

\begin{table*}[t]
\caption{Definitions and Descriptions of DFL Taxonomy}
\label{Table DFL Taxonomy}
\centering
\begin{tabularx}{\linewidth}{@{}l|l|X@{}}
\toprule
\textbf{Taxonomy} & \textbf{Category} & \textbf{Description} \\
\midrule
\multirow[t]{4}{*}{Iteration Order}
  & Sequential & Clients are synchronized to communicate one by one in a certain order. \\
~ & Random & Clients are synchronized to communicate one by one in a random order. \\
~ & Cycle & Clients are synchronized to communicate one by one in cycle. \\
~ & Parallel & All clients communicate asynchronously. \\
\midrule
\multirow[t]{3}{*}{Communication Protocol}
  & Pointing & Clients communicate in a specific form of one-peer-to-one-peer. \\
~ & Gossip & Clients communicate in a random form of one-peer-to-one-peer, which may be determined by the neighborhood principle, client model version, complete randomness, etc. \\
~ & Broadcast & Clients communicate in a form of one-peer-to-all-peers. \\
~ & Broadcast-gossip & Clients communicate in a form of one-peer-to-multiple-peers, which is also a combination form of Gossip and Broadcast. \\
\midrule
\multirow[t]{7}{*}{Network Topology}
  & Line & Clients communicate in a sequential pointing form. \\
~ & Bus & Clients send the model to all clients behind them in order. \\
~ & Ring & Clients communicate in a cycle pointing form. \\
~ & Mesh & Clients communicate with all other clients. \\
~ & Star & Clients communicate only with the central client. \\
~ & Tree & Clients communicate only with their central clients. \\
~ & Hybrid & Combination of multiple communication topologies. \\
\midrule
\multirow[t]{2}{*}{Paradigm Proposal}
  & Continual & Client learns directly from the model of the previous client. \\
~ & Aggregate & Client first aggregates the models of past clients and then learns on the aggregated models. \\
\midrule
\multirow[t]{2}{*}{Temporal Variability}
  & Static & Communication architecture will not change. \\
~ & Dynamic & Communication architecture may change with external factors, resource-saving purpose, fairness purpose, concept drift, etc. \\
\bottomrule
\end{tabularx}
\end{table*}

\section{Taxonomy of \\ Decentralized Federated Learning}
\label{Sec. Decentralized FL}

In this section, we begin by analyzing and comparing DFL and other related designations. Subsequently, we provide a well-organized, clear, and precise description of the various iterations, protocols, network topologies, paradigms, and variations in DFL, as presented in Table \ref{Table DFL Taxonomy}. It is worth noting that the table comprises five distinct taxonomies, which may exhibit overlapping meanings as well as conflicting aspects, and can also be applied in a complementary manner. These taxonomies, representing the viewpoints of the authors, include summarizations of existing literature, extensions of understanding, and even inferences regarding potential definitions. This comprehensive approach aims to strengthen the comprehension and categorization of concepts in the field of DFL.

\subsection{Iteration Order}
\label{Sec. Iteration Order}
In general, FL requires multiple iterations to converge, and iteration order represents the order of each client in each iteration or the way client queues are formed in DFL. In CFL, clients iterate in a parallel manner, and the order in which the server receives the client models does not affect the convergence of the system. However, in DFL, the iteration order of clients will significantly affect the performance of client models, and we continue to discuss this issue in depth in Section \ref{Sec. Paradigm Proposal}. Depending on the specific usage scenario and task requirements, the client iteration order in DFL can be determined to be sequential, cyclic, random, parallel, dynamic, or other strategies. The choice of iteration order can impact the convergence and performance of the system, and it is important to consider the specific characteristics and constraints of the application when determining the appropriate order.

\subsection{Communication Protocol}
\label{Sec. Communication Protocol}
DFL is a network framework for sharing model weights based on the pointing, gossip, or broadcast protocol, with the goal of obtaining optimal models across all clients. Pointing is one of the simplest and most straightforward forms of establishing a communication relationship between two peers in a unidirectional, one-to-one, and specified form. The algorithms of gossip and broadcast have been well established for use in networks \cite{nedic2018network}. Gossip protocol is essentially a random one-peer-to-one-peer way for clients to share, disseminate, and learn knowledge in a stochastic communication method \cite{boyd2006randomized,kempe2003gossip}. It is a standard communication protocol in DFL and is already in its infancy \cite{koloskova2019decentralized,hu2019decentralized}. The broadcast protocol is a one-peer-to-all-peers approach that allows the client to broadcast its model to all clients \cite{nedic2010asynchronous}. 

Hybrid protocols are now more popular, with different gossip, broadcast, and their combined communication structures designed for different scenarios and constraints. Aysal \textit{et al.} \cite{aysal2009broadcast} proposed a method that combines gossip and broadcast protocols and can be considered as a one-peer-to-neighbor-peers approach, where the client first broadcasts to its neighbors before gossiping. Bellet \textit{et al.} \cite{bellet2018personalized} introduced an algorithm that operates the agent asynchronously and performs broadcast communication between similar clients with a focus on obtaining personalized local models.

\subsection{Network Topology}
\label{Sec. Network Topology}
\begin{figure}[t]
\centering
\centerline{\includegraphics[width=\linewidth]{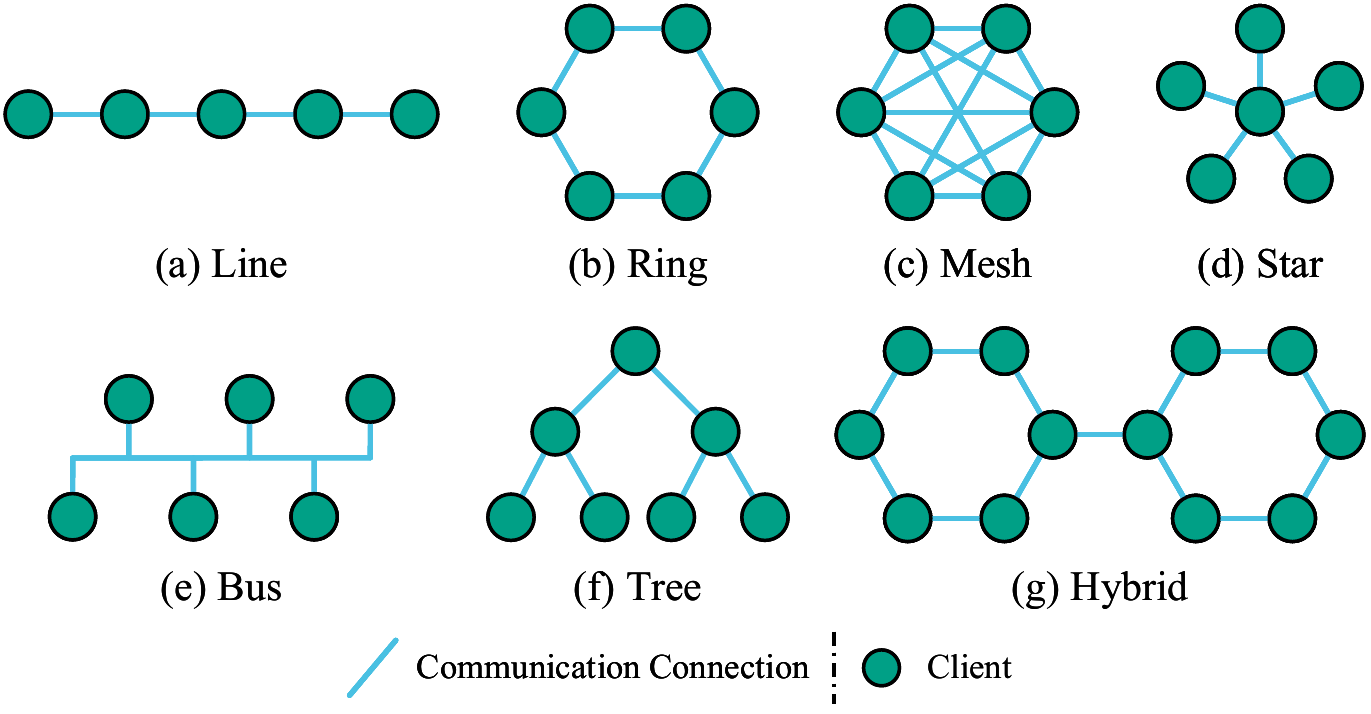}}
\caption{Illustration of communication network topology.}
\label{Fig. Network topology}
\end{figure}

DFL networks are inspired by various network topologies, as detailed in sources such as \cite{wang2019impact,neglia2019role,marfoq2020throughput,malandrino2021federated,chellapandi2024decentralized}. Nedić \textit{et al.} \cite{nedic2018network} highlighted and provided convergence proofs for several network structures, including grid, star, and fully connected topologies. Due to the absence of server-based adaptation, management, and propagation constraints, DFL networks exhibit a diverse range of configurations, as illustrated in Fig. \ref{Fig. Network topology}. Drawing from Graph FL concepts, DFL networks can form a graph to represent structured relationships, with clients as nodes and their connections as edges. This graph-based approach in DFL enables the quantification of dependencies between clients based on characteristics such as heterogeneity and communication patterns, thus enhancing learning efficiency and model robustness. Note that the depicted line segment indicates a connection between clients, which can be either unidirectional or bidirectional. The data transmitted between clients do not only comprise their models, but may also include models from previous interactions. In addition, the computational scope of each client can encompass both local learning and aggregation.

For CFL, since all client models are hosted on the server, aggregation (i.e., averaging of all model parameters) is the mainstream and popular method for integrating knowledge from all clients. However, for DFL, the situation is much more complex. First, the network topology is diverse. There are diverse network topologies in DFL. At each communication round, clients may obey different protocols to transmit models to one or more other clients. Second, there are different versions of the model. Except for the synchronous DFL, there must be different versions of the model for other DFLs. The subsequent clients in the learning process will have models that incorporate more knowledge compared to the previous clients. Third, acquiring all client knowledge becomes more challenging. Without a centralized server for collaborative management, future clients face difficulties in accessing the knowledge of all previous clients, except for the immediate preceding clients, such as Fig. \ref{Fig. Network topology}(a). Therefore, there is an urgent need for an alternative paradigm to complement and expand the FL landscape that is not well-compatible with aggregation.

\begin{table*}[t]
\caption{Paradigms of DFL}
\label{Table Paradigms of DFL}
\centering
\begin{tabularx}{\linewidth}{@{}p{50pt}|X|X@{}}
\toprule
\textbf{Paradigm} & \textbf{\texttt{Continual}} & \textbf{\texttt{Aggregate}} \\
\midrule
Intrinsic & 
Client receives a single model per iteration. \par Learning is performed on the received model without aggregation. & 
Client receives multiple models per iteration. \par Learning is performed locally after aggregating the previous models. \\
\midrule
Algorithm & 
\textbf{For} each client \textbf{until convergence do:}
\begin{enumerate} \item Receive the model from the previous client. \item Perform local learning on the model. \item Transmit the trained model to the next client. \end{enumerate} 
& \textbf{For} each client \textbf{until convergence do:}
\begin{enumerate} \item Receive all other client models from the previous client (pointing and gossip) or other clients (broadcast and broadcast-gossip). \item Aggregate all received models and perform local learning on the aggregated model. \item Transmit the trained model and other client models to the next client (pointing and gossip) or transmit the trained model to all clients (broadcast and broadcast-gossip). \vspace{-5pt} \end{enumerate} \\
\midrule
\multirow[t]{2}{*}{Advantage} & 
\multicolumn{2}{p{450pt}}{\begin{tableitemize} \item Each client involved in learning has a highly accurate, personalized, high-confidence local model. \item Compared to CFL, they do not have the same set of issues on the server side, such as aggregation fairness. \end{tableitemize}} \\
& \begin{tableitemize} \item Fewer communication, computation, and storage resources are required. \item More simple and straightforward, suitable for all scenarios. \end{tableitemize} 
& \begin{tableitemize} \item More powerful generalization ability on the obtained model. \item Stronger ability to update new knowledge generated by the client. \end{tableitemize} \\
\midrule
\multirow[t]{2}{*}{Challenge} & 
\multicolumn{2}{p{450pt}}{\begin{tableitemize} \item Model performance strongly depends on the client iteration order. \item Appropriate loss function, learning rate, and training epoch, which allows the model to learn the current client's knowledge while ensuring that the previous knowledge is not forgotten. \end{tableitemize}} \\
& \begin{tableitemize} \item Catastrophic forgetting of past client knowledge. \end{tableitemize}
& \begin{tableitemize} \item Repetition and overemphasis on learning from past clients. \end{tableitemize} \\
\midrule
Network Topology (Take sequential pointing line DFL architecture as an example) & \multicolumn{2}{p{450pt}}{\vspace{-8pt}\begin{minipage}[t]{\linewidth} \includegraphics[width=\linewidth]{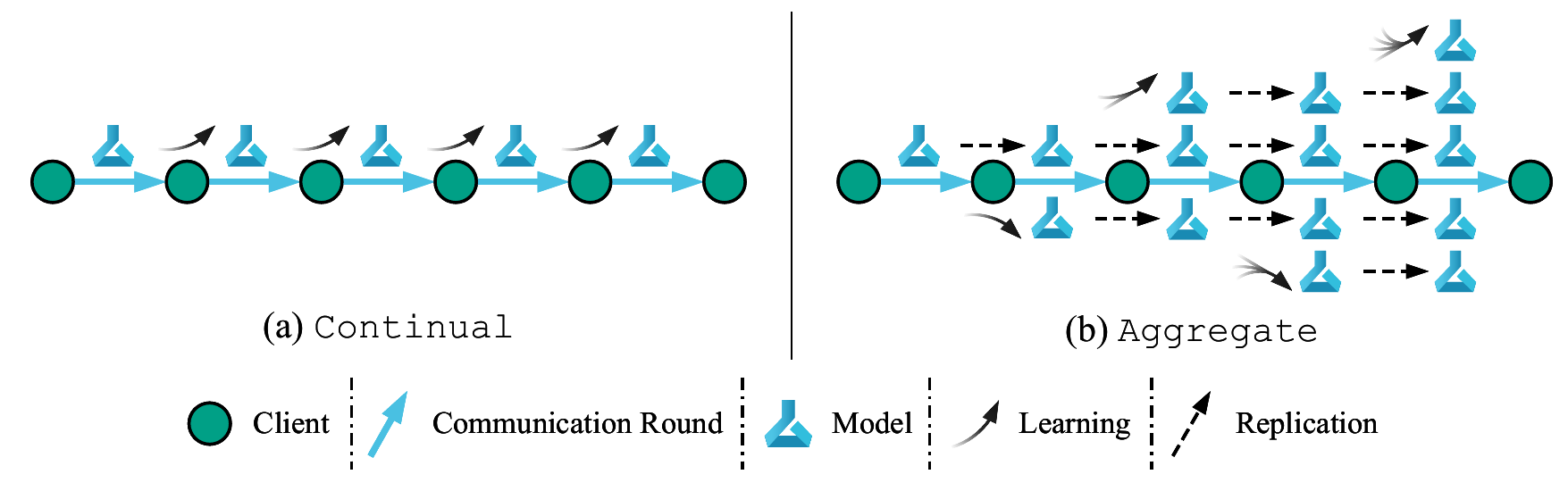} \end{minipage}} \\ 
\bottomrule
\end{tabularx}
\end{table*}

\subsection{Paradigm Proposal}
\label{Sec. Paradigm Proposal}
We introduce an innovative taxonomy of DFL into two paradigms: \texttt{Continual} and \texttt{Aggregate}. The main differences between these two paradigms lie in the number of model updates exchanged between clients and whether aggregation takes place. The distinction between the paradigms also entails variations in other settings, such as learning rates. The \texttt{Aggregate} paradigm represents the archetypal FL algorithm, where each client receives the model from other clients, aggregates these models, and subsequently conducts local learning. Conversely, within the \texttt{Continual} paradigm, each client receives the model from merely one peer client and proceeds to learn directly based upon this particular model. Continual learning \cite{lesort2020continual,delange2021continual}, or be called incremental learning, provides a solid and grounded theory for \texttt{Continual}. A number of concepts and algorithms for federated continual learning are mentioned in the recent literature \cite{criado2022non,usmanova2021distillation,park2021tackling,yoon2021federated}, which consider the process of dynamic data collection in the real world while addressing the issues of non-IID data, concept drift, and catastrophic forgetting. In DFL, the role of continual learning is more extensive:
\begin{itemize}
    \item The subsequent client will directly learn on the model of the previous client. Compared to local learning and \texttt{Aggregate}, the client is able to obtain a more personalized model while saving communication, computational, and storage resources.
    \item In storage-constrained frameworks, clients do not need to retain any additional model parameter data.
    \item In computation-constrained frameworks, clients also do not need to consume additional resources for aggregation calculations.
    \item The continuous generation of new data by clients is accommodated, and they do not need to wait for all data to be collected before starting the local learning process.
    \item For tasks that may have concept drift, clients are always provided with the latest version of the model.
\end{itemize}

Table \ref{Table Paradigms of DFL} analyzes and summarizes the intrinsic, algorithm, advantage, challenge, and network topology of these two paradigms. The difference between these two paradigms is illustrated by the example of sequential pointing line DFL. In the \texttt{Continual} paradigm, the only content delivered to the subsequent client is the trained model. The subsequent client just continue learning on this model, as shown in Table \ref{Table Paradigms of DFL}(a). In the \texttt{Aggregate} paradigm, the previous client transmits not only the trained local model but also all the previous models. The learning process performed by the subsequent client is divided into two parts, first aggregation and then learning, as shown in Table \ref{Table Paradigms of DFL}(b). 

In order to compare and illustrate the difference between the \texttt{Continual} and \texttt{Aggregate} paradigms, pointing, gossip, and broadcast, and different network topologies. Algorithm \ref{Alg line} shows two paradigms in the sequential pointing line DFL topology and Algorithm \ref{Alg pointing ring and broadcast mesh} shows pointing ring \texttt{Continual} and {broadcast mesh \texttt{Aggregate} DFL. The difference between the \texttt{Continual} and \texttt{Aggregate} paradigms can be clearly seen in the pre-processing of the client before learning and the sharing of the model after learning. The additional requirements of the \texttt{Aggregate} paradigm for communication, computation, and storage have been highlighted. Under the \texttt{Aggregate} paradigm, the pointing and gossip protocols require the client to send more model data at once, while the broadcast and broadcast-gossip protocols require the client to send at a higher frequency. The ring topology can be seen as a cyclic variant of the line topology, and both network topologies are widely used by researchers due to their simple and straightforward structure. The line topology is a sufficient knowledge learning system for systems that do not generate new knowledge. However, in a system that is constantly generating new knowledge, the ring topology may be a more reasonable topology. It is not only able to re-update the knowledge in the system but also a feasible solution to catastrophic forgetting. 

To further illustrate the learning and communication process among clients in these two paradigms, Fig. \ref{Fig. Parameter Space} demonstrates the learning process from the first client to the final client in the parameter space. It is important to note that we actually have several assumptions here. Firstly, the optimal solutions of the local models of all clients follow a multivariate normal distribution in the parameter space. Secondly, considering the systemic and statistical heterogeneity, some clients exhibit significant biases. Thirdly, although reasonable loss functions and learning rates are chosen, the models are not always trained to achieve optimal solutions. The communication and learning processes of the two paradigms, \texttt{Continual} and \texttt{Aggregate}, are as follows.
\begin{enumerate}[label=Step \arabic*),leftmargin=35pt]
\item Both paradigms initiate learning with the same initial model parameters and obtain the same Model 1 in Client 1.
\item Both paradigms learn from Model 1 and reach the same Model 2 in Client 2. It's worth noting that the \texttt{Aggregate} paradigm is meaningful when there are two or more aggregated models available.
\item In the \texttt{Continual} paradigm, Client 3 learns directly from Model 2 to obtain Model 3, while in the \texttt{Aggregate} paradigm, Model 1 and Model 2 are first aggregated, and then Client 3 learns from the aggregated model to obtain Model 3. Note that the \texttt{Continual} paradigm is less complex than the \texttt{Aggregate} paradigm, as indicated by the length of the black arrow.
\item In the \texttt{Aggregate} paradigm, the model aggregated by Client 4 is closer to the center of the Normal distribution than Client 3, so it is expected that the learning process for subsequent clients will be easier.
\item [Step $n$)] When the client is positioned towards the end of the queue, the learning difficulty in the \texttt{Continual} paradigm becomes random, depending on the deviation between the previous client and the current client. However, in the \texttt{Aggregate} paradigm, the learning difficulty is only influenced by the current client since the aggregated model is expected to be extremely close to the center of the normal distribution.
\end{enumerate}

\begin{algorithm}[t]
\small
   \caption{\small{Sequential \colorbox{SpringGreen}{pointing line \texttt{Continual}} and \colorbox{Thistle}{pointing line \texttt{Aggregate}} decentralized federated learning.}}
   \label{Alg line}
\begin{algorithmic}
   \State {\bfseries Input:} Client set ($C$), training epoch ($E$), initial model ($\omega_0$), loss function ($\mathcal{L}$), learning rate ($\eta$)
   \State {\bfseries Output:} Local models ($\{\omega_{c} | c \in C\}$)
    \For{$c \in C$ {\bfseries in sequence}}
        \State \colorbox{SpringGreen}{Copy the model from previous client $\omega_c \gets \omega_{c-1}$}
        \State \colorbox{Thistle}{Aggregate received models $\omega_{c} \gets \text{Aggregation}\{\omega_{1}, \omega_{2}, ..., \omega_{c-1}\}$}
        \For{$e=1$ {\bfseries to} $E-1$}
            \State Backpropagate and update the local model $\omega_c^{e+1} \gets \omega_c^{e} - \eta\nabla\mathcal{L} $.
        \EndFor
        \State Update the local model $\omega_c \gets \omega_c^{E}$.
        \State Client $c$ sends \colorbox{Thistle}{$\{\omega_{1}, \omega_{2}, ..., \omega_{c-1}\}$ and} $\omega_c$ to the next client.
        \State 
    \EndFor
\end{algorithmic}
\end{algorithm}

Based on the aforementioned assumptions and iterative process, we can make certain expectations regarding the accuracy, loss, convergence, and communication complexity of the clients in both paradigms during training. We come up with the following speculations: 
\begin{enumerate}
    \item The learning loss will exhibit periodic oscillations across client iterations and eventually converge in both paradigms.
    \item The convergence of learning loss in the \texttt{Aggregate} paradigm is expected to be more stable. In the \texttt{Continual} paradigm, the learning difficulty depends on the discrepancy between the previous client and the current client's local data, in other words, it depends on the iteration order of clients. Under the assumption of a normal distribution, the learning difficulty in the \texttt{Aggregate} paradigm is determined by the heterogeneity of the current client's data, and most clients may have similar data distributions.
    \item The convergence of learning loss in the \texttt{Aggregate} paradigm is also expected to be faster due to the decreasing learning difficulty as the client iterations progress. However, this acceleration in convergence is accompanied by an increase in communication overhead.
    \item The \texttt{Continual} paradigm requires more communication rounds to achieve convergence, whereas the \texttt{Aggregate} paradigm incurs greater communication overhead per round.
    \item The \texttt{Continual} paradigm exhibits stronger personalization, while the \texttt{Aggregate} paradigm demonstrates greater generalization. Depending on scenario requirements, both paradigms can achieve similar performance after convergence by adjusting weights.
\end{enumerate}

\begin{algorithm}[t]
\small
   \caption{\small{Cycle \colorbox{SpringGreen}{pointing ring \texttt{Continual}} and \colorbox{Thistle}{broadcast mesh \texttt{Aggregate}} Decentralized Federated Learning.}}
   \label{Alg pointing ring and broadcast mesh}
\begin{algorithmic}
   \State {\bfseries Input:} Client set ($C$), training epoch ($E$), initial model ($\omega_0$), loss function ($\mathcal{L}$), learning rate ($\eta$)
   \State {\bfseries Output:} Local models ($\{\omega_{c} | c \in C\}$)
    \While{$c \in C$ {\bfseries in cyclic}} \Comment{line vs. ring}
        \State \colorbox{SpringGreen}{Copy the model from previous client $\omega_c \gets \omega_{c-1}$}
        \State \colorbox{Thistle}{Aggregate received models $\omega_{c} \gets \text{Aggregation}\{\omega_{1}, \omega_{2}, ..., \omega_{c-1}\}$}
        \For{$e=1$ {\bfseries to} $E-1$}
            \State Backpropagate and update the local model $\omega_c^{e+1} \gets \omega_c^{e} - \eta\nabla\mathcal{L} $.
        \EndFor
        \State Update the local model $\omega_c \gets \omega_c^{E}$.
        \State Client $c$ sends $\omega_c$ to the next client \colorbox{Thistle}{and all other clients}. 
        \State \Comment{pointing vs. broadcast} 
    \EndWhile
\end{algorithmic}
\end{algorithm}

\subsection{Temporal Variability}
\label{Sec. Temporal Variability}
The network topology of DFLs has recently undergone a shift from static to dynamic trends, adapting to the time-varying external environment \cite{Parasnis2023ConnectivityAwareSF}. The inspiration for the separation and clustering of network topologies comes from group behaviors observed in nature, such as fish schools and bee swarms \cite{sayed2013diffusion}. When a school of fish encounters a predator, the entire school separates to avoid it. Similarly, in a bee swarm, a small number of scouts can lead the entire swarm, demonstrating the herd effect. Interestingly, migratory birds form V-shaped formations during long-distance flights to conserve energy, and the birds at the front rotate over time to distribute flight fatigue evenly. In the context of DFL networks, dynamic topologies may exhibit more robust, fair, and efficient performance compared to static topologies. The determination of dynamic topologies in DFL networks can be influenced by various factors, including:
\begin{itemize}
    \item \textbf{External Interference.} Strong and unbreakable communication barriers, SPoF, malicious attacks, and other external factors can lead to changes in the network topology. In order to avoid the failure of the entire network, topology adjustments and discards are made.
    \item \textbf{Communication Resource Saving.} Clients have the ability to dynamically select their neighbors for each communication. By selectively choosing nearby clients, communication resources can be optimized and saved. Additionally, clients can dynamically elect the most central client as a leader during each communication, enhancing the efficiency and effectiveness of communication within the network.
    \item \textbf{Fairness.} In order to ensure fairness among clients, a random selection process is employed for determining the communication target. This helps to prevent any bias or preference towards specific clients, ensuring equal opportunities for all participants.
\end{itemize}
The development of dynamic topological structures based on these factors shapes DFL networks to facilitate robust, efficient, and fair communication among clients.

\section{Variants of Decentralized Federated Learning}
\label{Sec. Variants of DFL}
In Section \ref{Sec. Decentralized FL}, we provide a comprehensive definition, introduction, and propose two paradigm for DFLs. In this section, we review the real-world applications in DFL, with a specific focus on its diverse applications across various domains and its real-world deployment. Taking inspiration from the CFL variants and considering the underlying network topologies depicted in Fig. \ref{Fig. Network topology}, we propose several viable topology variants for DFL. These topology variants serve as alternative options for researchers to consider when deploying DFL. We discuss the advantages and limitations associated with each variant, enabling researchers to make informed decisions regarding the most suitable topology for specific usage scenarios.

\begin{figure*}[t]
\centering
\centerline{\includegraphics[width=\linewidth]{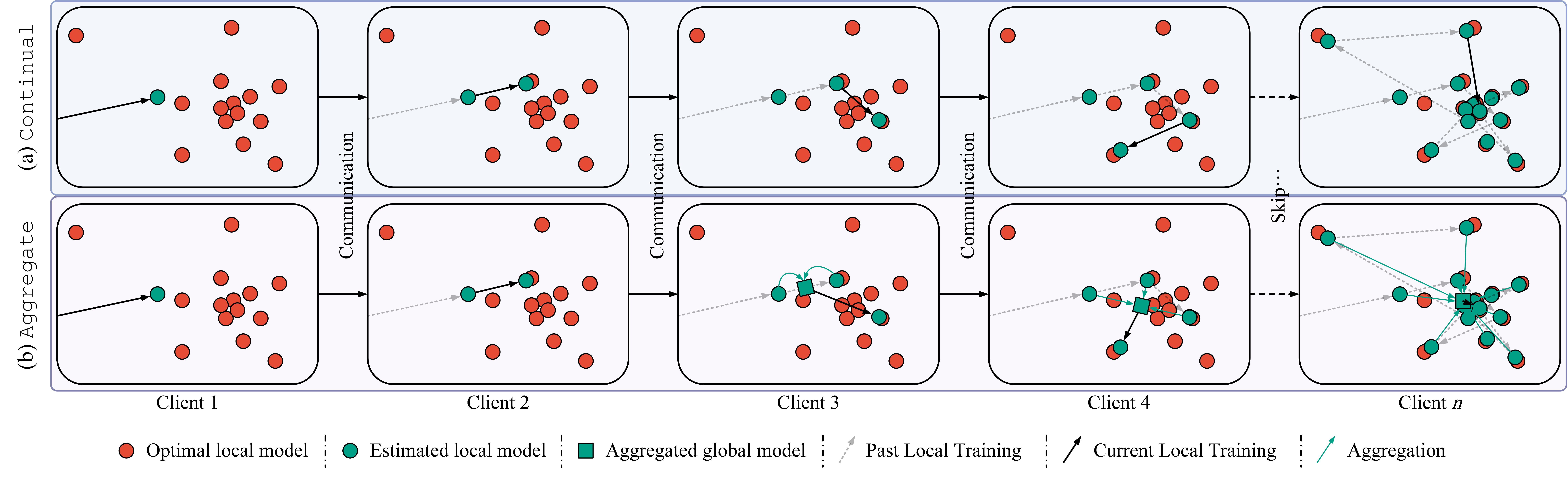}}
\caption{Illustration of the two paradigms, \texttt{Continual} and \texttt{Aggregate}, for sequential pointing line DFL in the parameter space, showcasing their respective learning and communication processes. The length of the arrow represents both the learning difficulty and the magnitude of the model parameters that undergo changes during learning, which can be measured using the $\ell_2$ norm. Shorter arrows are desired as they indicate more accessible, stable, and accurate model learning and convergence. Excessively long arrows suggest that the given loss function and learning rate may not produce the desired model outcome.}
\label{Fig. Parameter Space}
\end{figure*}

\begin{table*}[t]
\renewcommand*{\arraystretch}{1.5}
\caption{Some Inspiring DFLs with Different Protocols, Topologies, Paradigms, and Variants.}
\label{Table inspiring DFLs}
\centering
\begin{tabularx}{\linewidth}{lllp{90pt}X}
\toprule
\textbf{Literature} & \textbf{Year} & \textbf{Paradigm} & \textbf{Type} & \textbf{Highlight} \\
\specialrule{\lightrulewidth}{0.2 em}{0pt}
\rowcolor{Gray}
Chang \textit{et al.} \cite{chang2018distributed} & 2018 & \texttt{Continual} & \begin{tableitemize} \item Sequential pointing line \item Cycle pointing ring \end{tableitemize} & \begin{tableitemize} \item Introduced system heterogeneity artificially. \end{tableitemize} \\
Sheller \textit{et al.} \cite{sheller2019multi} & 2019 & \texttt{Continual} & \begin{tableitemize} \item Sequential pointing line \item Cycle pointing ring \end{tableitemize} & \begin{tableitemize} \item Obtained a conclusion that catastrophic forgetting worsens as the number of clients increases. \end{tableitemize} \\
\rowcolor{Gray}
Sheller \textit{et al.} \cite{sheller2020federated} & 2020 & \texttt{Continual} & \begin{tableitemize} \item Sequential pointing line \item Cycle pointing ring \end{tableitemize} & \begin{tableitemize} \item Considered the DFL framework to output a final model approach. \end{tableitemize} \\
Huang \textit{et al.} \cite{huang2022continual} & 2022 & \texttt{Continual} & \begin{tableitemize} \item Sequential pointing line \item Cycle pointing ring \end{tableitemize} & \begin{tableitemize} \item Introduced synaptic intelligence in \texttt{Continual} DFL to effectively improve model stability, especially for sequential pointing line topology. \end{tableitemize} \\
\rowcolor{Gray}
Yuan \textit{et al.} \cite{yuan2023peer} & 2023 & \texttt{Continual} & \begin{tableitemize} \item Random gossip ring \end{tableitemize} & \begin{tableitemize} \item Considered the highly dynamic and random nature of vehicle connectivity in vehicular networks and employed gossip-based communication to simulate this characteristic when deploying \texttt{Continual} DFL. \item Provided a comprehensive comparison between CFL and DFL, such as knowledge dissemination mechanism, communication complexity, generalizability, compatibility, overhead, hidden concerns, etc. \end{tableitemize} \\
Assran \textit{et al.} \cite{assran2019stochastic} & 2019 & \texttt{Aggregate} & \begin{tableitemize} \item Cycle broadcast-gossip mesh \item Parallel broadcast mesh \end{tableitemize} & \begin{tableitemize} \item Performed a comparison of broadcast-gossip and broadcast protocols. \end{tableitemize} \\
\rowcolor{Gray}
Roy \textit{et al.} \cite{roy2019braintorrent} & 2019 & \texttt{Aggregate} & \begin{tableitemize} \item Random broadcast-gossip mesh \end{tableitemize} & \begin{tableitemize} \item Pre-requested model versions from other clients. \item Considered the scenario where the DFL framework outputs a model to a new client. \end{tableitemize} \\
Pappas \textit{et al.} \cite{pappas2021ipls} & 2021 & \texttt{Aggregate} & \begin{tableitemize} \item Parallel broadcast star \end{tableitemize} & \begin{tableitemize} \item Proposed a framework that combines star DFL with split learning. \end{tableitemize} \\
\rowcolor{Gray}
Warnat \textit{et al.} \cite{warnat2021swarm} & 2021 & \texttt{Aggregate} & \begin{tableitemize} \item Dynamic pointing star \end{tableitemize} & \begin{tableitemize} \item Elected a leader dynamically via a blockchain smart contract that is used to aggregate model parameters. \end{tableitemize} \\
Shi \textit{et al.} \cite{shi2021over} & 2021 & \texttt{Aggregate} & \begin{tableitemize} \item Cycle broadcast-gossip hybrid \end{tableitemize} & \begin{tableitemize} \item Analyzed the convergence of broadcast-gossip in a hybrid network. \end{tableitemize} \\
\rowcolor{Gray}
Chen \textit{et al.} \cite{chen2022decentralized} & 2022 & \texttt{Aggregate} & \begin{tableitemize} \item Cycle broadcast mesh \end{tableitemize} & \begin{tableitemize} \item Introduced the superposition property of the analog scheme to improve the parallelism of communication, which enables a significant reduction communication rounds. \end{tableitemize} \\
Wang \textit{et al.} \cite{wang2022matcha} & 2022 & \texttt{Aggregate} & \begin{tableitemize} \item Dynamic parallel broadcast-gossip hybrid \end{tableitemize} & \begin{tableitemize} \item Promoted more frequent communication in the central client to achieve fast convergence. \item Promoted less frequent communication in the other clients to achieve low communication latency. \item The proposed algorithm is applicable and generalized to all hybrid networks. \end{tableitemize} \\
\bottomrule
\end{tabularx}
\end{table*}

\subsection{Real-World Applications}
\label{Sec. Real-World Applications}

The development of a DFL framework relies on several key factors, such as relevant application scenarios, sources of information acquisition, information processing units, and perceptual prediction modules, among others. With the establishment of the theoretical framework for networks, DFL has been adopted in various application domains, including vehicles, healthcare, industrial IoT, social networks, etc.}

\textit{1) \textbf{Connected and Automated Vehicles (CAVs)}} serve as a robust hardware infrastructure for DFL, leveraging onboard batteries, diverse sensors, computing units, storage devices, and more. Existing vehicle networking frameworks, such as vehicle-to-vehicle (V2V), have also laid the foundation for communication and networking experiences in DFL for CAV \cite{harding2014vehicle,chellapandi2023federated}. Referred to as V2V FL, this approach enables the exchange and sharing of up-to-date knowledge among vehicles and has been explored in recent studies \cite{samarakoon2019distributed,du2020federated,pokhrel2020decentralized,yu2020proactive,chen2021bdfl,barbieri2022decentralized,nguyen2022deep,su2022boost}. Lu \textit{et al.} \cite{lu2020federated} proposed a vehicular DFL approach with a focus on privacy protection and mitigating data leakage risks in vehicular cyber-physical systems (VCPS). In their framework, roadside units (RSUs) are responsible for forwarding vehicle identities, vehicle data retrieval information, data profiles, data sharing requests, and related tasks. Once the V2V connection is established through the RSU intermediary, the model data is directly transmitted to the requesting vehicle.

\textit{2) \textbf{Healthcare Institutions}} are inclined towards DFL frameworks over CFL due to their abundant patient privacy data, computational resources, and storage capabilities. As key stakeholders in healthcare institutions, clinicians play a crucial role in data collection, model training, data analysis, characterization, and providing experimental results and solutions. Unlike traditional server-centric approaches, clinicians have the flexibility to observe, analyze, fine-tune, and match models manually, offering more control and adaptability. Healthcare institutions widely employ DFL frameworks in various studies \cite{chang2018distributed,sheller2019multi,sheller2020federated,huang2022continual,xu2021federated,tedeschini2022decentralized,nguyen2022novel,baghersalimi2023decentralized}. Warnat-Herresthal \textit{et al.} \cite{warnat2021swarm} introduced a DFL framework called Swarm Learning, which addresses four use cases of heterogeneous diseases, including COVID-19, tuberculosis, leukemia, and lung pathologies. This framework incorporates a blockchain smart contract for enhanced security and dynamically selects a leader for aggregating model parameters in each iteration.

\textit{3) \textbf{Industrial IoT (IIoT)}}, as a cornerstone of Industry 4.0, greatly benefits from DFL, which offers robust and scalable solutions tailored to the complexities of modern manufacturing environments \cite{qu2020blockchained,ranathunga2022blockchain,friha20232df}. IIoT requires enhanced robustness, and DFL improves system resistance to SPoF, which is particularly crucial for devices on production lines \cite{qiu2022decentralized}. The high autonomy of DFL makes it well suited for industrial settings with diverse physical conditions and operational environments. Moreover, IIoT can rely on the scalability of DFL to easily adapt to geographically dispersed production sites and dynamically accommodate new industrial devices into the DFL system. Du \textit{et al.} \cite{du2022decentralized} designed a DFL framework for IIoT, where each client exchanges model parameters only with neighbors using a broadcast-gossip communication protocol to achieve model consensus. To enhance the efficiency of the gossip protocol and reduce communication overhead, they also consider the topology of the entire client network to facilitate asynchronous model exchanges between clients.

\textit{4) \textbf{Mobile Services}} based on IoT devices provide a significant application scenario for DFL, leveraging the capabilities of smartphones, laptops, tablets, etc. These mobile IoT devices are equipped with various sensors, such as global positioning system (GPS), inertial measurement unit (IMU), cameras, sound sensors, and magnetic sensors, enabling them to acquire diverse sources of information. Unlike the relatively fixed connectivity of CAVs, mobile IoT devices offer more flexible systems and platforms to support a wide range of applications. In recent studies, DFL frameworks have been developed specifically for mobile IoT devices, aiming to leverage their computational power and sensor capabilities \cite{wilhelmi2022decentralization,koloskova2019decentralizedmobile}. While the traditional example of CFL, such as Google mobile keyboard prediction, is well-known \cite{hard2018federated}, the transfer of such applications to DFLs is of great interest. For instance, building DFLs among individuals with similar professions, such as doctors, lawyers, or engineers, can enable personalized word recommendations tailored to their specific needs. Belal \textit{et al.} \cite{belal2022pepper} developed a smartphone-based DFL personalized recommendation system for New York City attractions and movies. By sharing model parameters with neighbors who have similar interests, the system achieves higher hit rates and faster convergence, enhancing the recommendation accuracy and user experience.

\textit{5) \textbf{UAVs and Satellites}}, operating in dynamic computing environments, possess vast amounts of sensitive data for remote sensing, target recognition, and military-related tasks, under the constraints of limited resources, making them well suited for the advantages of DFL \cite{xiao2021fully,yan2023convergence,zhai2023fedleo}. For mobile UAVs and satellites, bandwidth is a valuable resource. One potential solution is that using a DFL framework based on broadcast gossip tailored to dynamic geographic locations can significantly reduce bandwidth requirements and the consumption of communication resources \cite{qu2021decentralized}. Moreover, due to their dynamic nature and highly variable data, DFL can also enhance real-time responsiveness, adaptability, and efficiency of task execution. Han \textit{et al.} \cite{han2024cooperative} proposed a DFL framework that orchestrates satellite constellations, aggregating models within a satellite cluster and relaying models to other satellites via inter-satellite links, particularly considering the dynamic scenarios of low earth orbit satellites with varying orbits.

\textit{6) \textbf{Social Networks}}, as large-scale knowledge graphs, contain various users as nodes, content, and connections, making DFL highly effective for handling a widely dispersed and personalized user base, where each user is connected only to their neighbors \cite{he2022spreadgnn}. Use cases in social networks include sentiment analysis, recommendation systems, publication systems, and influence analysis, among others \cite{he2021fedgraphnn,han2022demystifying,che2022decentralized}. Users in social networks come from diverse backgrounds, such as teachers and doctors, but may share common interests in topics like comics, leading to frequent interactions. The connections within and between user groups vary in proximity, thus emphasizing the need for a personalized and scalable DFL approach. Chen \textit{et al.} \cite{chen2023dfedsn} developed a DFL framework for social networks that establishes a user data structure with affine distributions. This structure helps capture the heterogeneity among users and reduces the loss associated with independently and identically distributed data.

\textit{7) \textbf{Artificial General Intelligence (AGI)}}, as one of the popular research areas today, represented by large language models (LLMs) like ChatGPT, has the potential to fundamentally change our lives \cite{achiam2023gpt}. The training of these models relies on vast amounts of computational resources and data, which can benefit from the collaborative learning paradigm of DFL to facilitate cooperation among different data centers \cite{kersic2024review,su2024towards}. Unlike other application areas discussed earlier, such as healthcare, AGI models feature billions or even tens of billions of model parameters and are intended for general use cases, not just confined to a specific application. This poses unique challenges for the application of DFL in AGI. Qin \textit{et al.} \cite{qin2023federated} proposed a method using FL for full-parameter fine-tuning of billion-scale LLMs, employing zeroth-order optimization with a random seed subset, reducing communication requirements to just a few random seeds and scalar gradients, totaling only a few thousand bytes. Meanwhile, Gao \textit{et al.} \cite{gao2023gradientcoin} only proposed a purely theoretical design for decentralized LLMs. Therefore, the use of DFL for AGI remains a nascent field requiring further research \cite{chen2023challenges}.


\begin{figure*}[t]
\footnotesize
\centering
\subfloat[Line/Ring]{\includegraphics[width=0.5\linewidth]{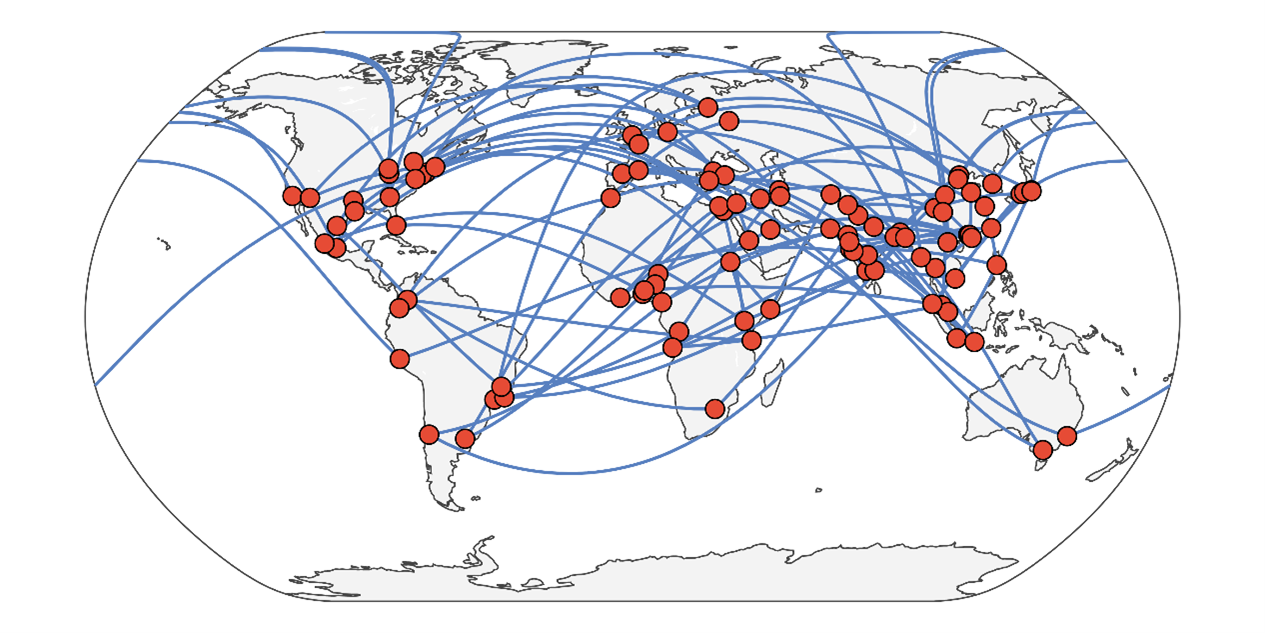}}
\hfill
\subfloat[Mesh]{\includegraphics[width=0.5\linewidth]{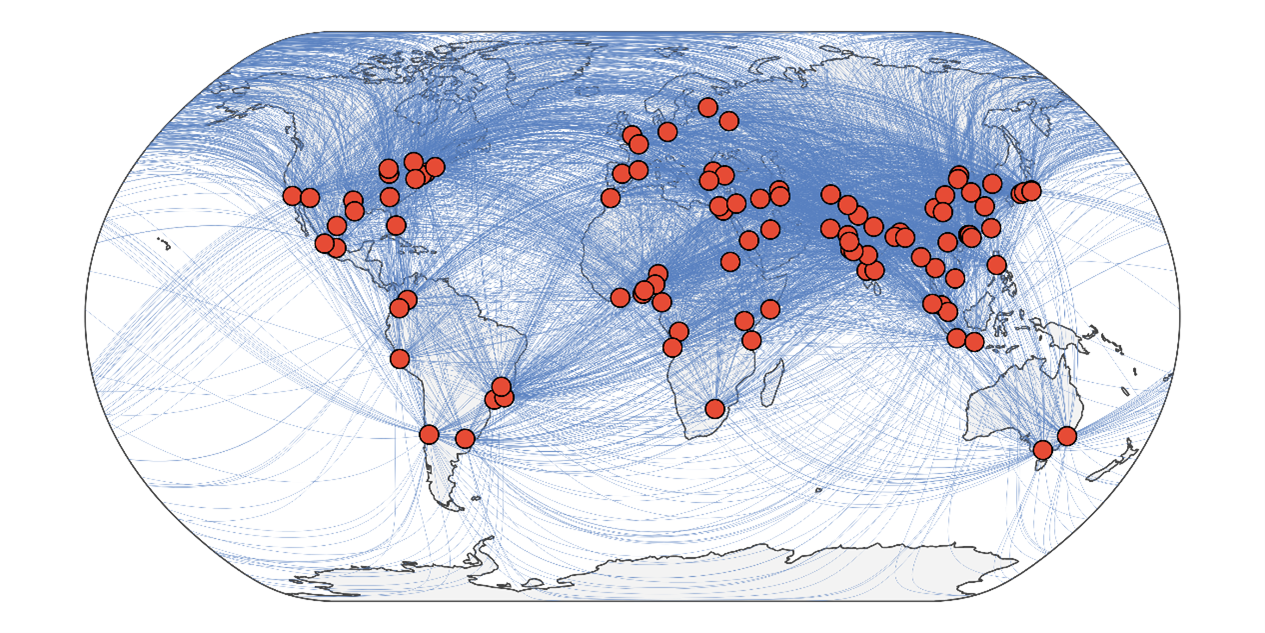}}
\\
\subfloat[Star]{\includegraphics[width=0.5\linewidth]{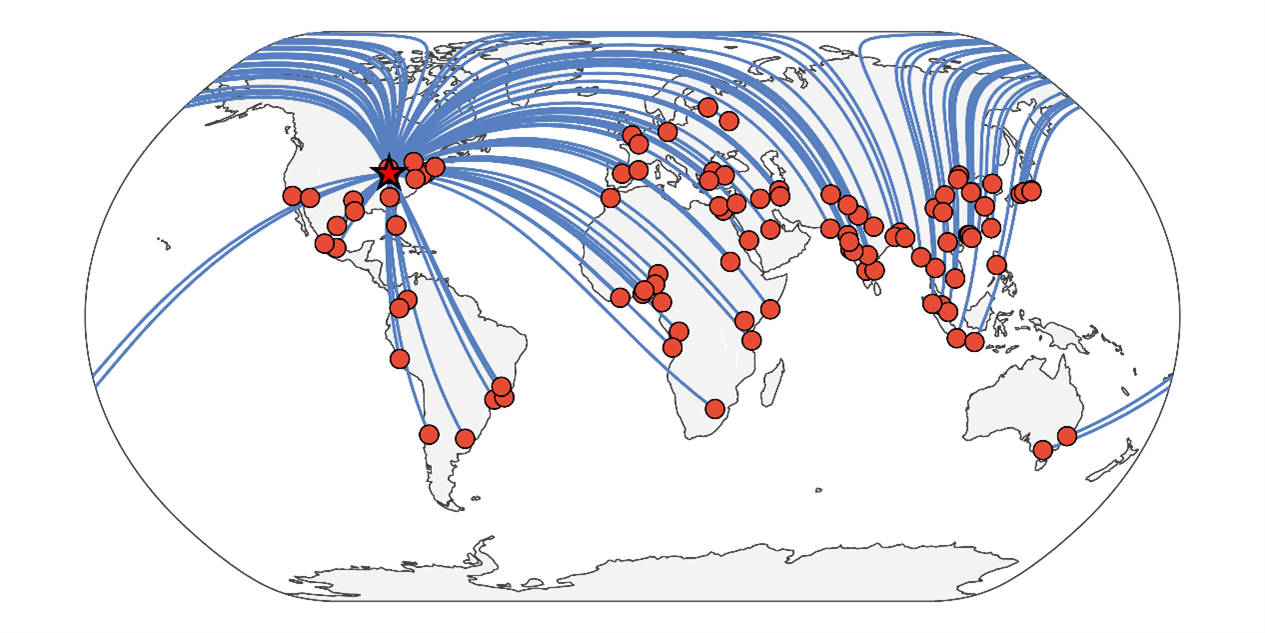}}
\hfill
\subfloat[Hybrid]{\includegraphics[width=0.5\linewidth]{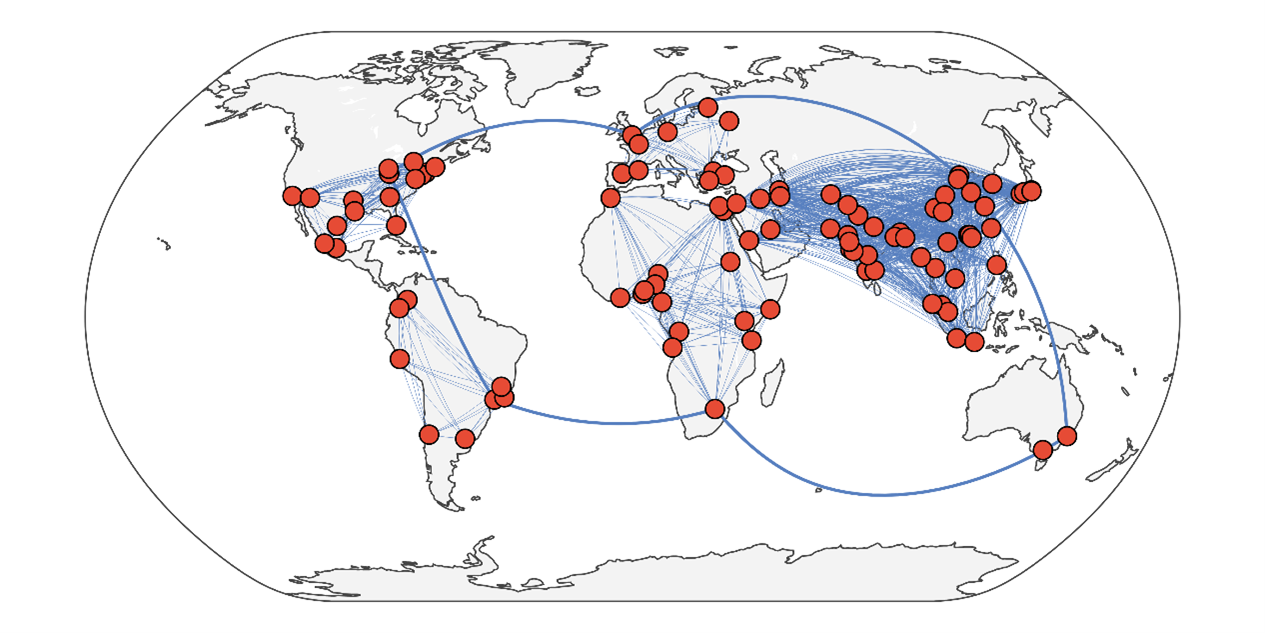}}
\caption{Illustrations of imagined DFL network topologies in the real world: (a) line/ring, (b) mesh, (c) star, and (d) hybrid. The red dots represent clients, which can be universities, institutions, or organizations in some of the major cities in the world (determined by population). The blue lines depict the communication network among these clients. Depending on the chosen topology, the communication networks exhibit different communication distances, number of communication links, complexity, and other characteristics.}
\label{Fig. World Network} 
\end{figure*}

\subsection{Variant: Line}
\label{Sec. Variant: Line}
The base variant of DFL can be considered as a sequential pointing line, depicted in Fig. \ref{Fig. Network topology}(a) and Fig. \ref{Fig. World Network}(a). This topology serves as the simplest and most straightforward illustration and comparison in this paper, as demonstrated in Table \ref{Table Paradigms of DFL}(a), (b), and Algorithm \ref{Alg line}. The line variant is frequently used as a baseline for comparison due to its ease of implementation, intuitiveness, and efficiency \cite{chang2018distributed,sheller2019multi,sheller2020federated,huang2022continual}. However, it has notable limitations, such as the inability to accommodate continuous learning of new knowledge within the system, the risk of catastrophic forgetting in the \texttt{Continual} paradigm, or redundant and excessive learning in the \texttt{Aggregate} paradigm, as well as limited generalization ability for starting clients and the vulnerability to a SPoF. Furthermore, the line variant lacks cyclic connections, limiting each client to a single iteration and preventing the system from fully converging. In particular, in the line variant, the clients at the front of the queue will have worse model performance. Given its prominent disadvantages and advantages, it can serve as a baseline or initial implementation for further research and development.

\subsection{Variant: Ring}
\label{Sec. Variant: Ring}
The ring variant corresponds to the cycle pointing line DFL, as depicted in Fig. \ref{Fig. Network topology}(b) and Fig. \ref{Fig. World Network}(a). The cyclic form is commonly used in DFL as the model needs to be trained between clients to acquire new knowledge collected from other clients, thereby enhancing generalization. Based on the framework, not all past models need to be transferred for aggregation in each communication since many of them may already be outdated. The ring variant not only inherits the simplicity of the line variant but also becomes a popular approach in various research papers due to its ability to iterate indefinitely until convergence.

The ring topology is already considered mature in decentralized learning \cite{lian2018asynchronous,koloskova2019decentralized} and is beginning to gain traction in DFL \cite{wang2022efficient}. For example, Chang \textit{et al.} \cite{chang2018distributed} proposed two heuristics for DFL, including sequential pointing communication on each client for one iteration and multiple iterations to obtain the final model. Similarly, Sheller \textit{et al.} \cite{sheller2020federated} also considered sequential pointing or cycle continual learning in the client to generate the final model. Nguyen \textit{et al.} \cite{nguyen2022deep} applied cycle pointing DFL to autonomous driving applications. Yuan \textit{et al.} \cite{yuan2023peer} proposed a random ring topology DFL framework, named FedPC, based on the gossip communication protocol for naturalistic driving action recognition. FedPC emphasizes the highly dynamic, random, and data-heterogeneous nature of vehicle connections in this context.

\subsection{Variant: Mesh}
\label{Sec. Variant: Mesh}
A multidirectional ring, also known as a fully connected topology, or be called mesh, is a variant of the ring basic variant, depicted in Fig. \ref{Fig. Network topology}(c) and Fig. \ref{Fig. World Network}(b). In the ring variant, each client needs to transmit multiple model parameters in each communication round, which can pose a burden on the network bandwidth. In contrast, the mesh variant requires each client to transmit its local model parameters to all other clients in each communication round. This approach entails higher communication frequency for the clients while also reducing the size of model packets transmitted per communication. The higher communication frequency and larger per-communication data packet overhead have their respective advantages and disadvantages, which can be traded off depending on the specific application context. However, when compared to the ring variant, the mesh variant significantly mitigates the impact of SPoF, which is a notable advantage of this variant.

Recent research has witnessed the emergence of mesh-based DFL approaches \cite{xu2021federated,wink2021approach,barbieri2022decentralized,chen2022decentralized}. Assran \textit{et al.} \cite{assran2019stochastic} proposed Stochastic Gradient Push (SGP), a parallel broadcast-gossip mesh DFL approach. In the broadcast-gossip iteration, clients in SGP send their trained local models to a sparse selection of other clients in a parallel manner, and they also receive models from other selected clients. Each client then performs a weighted aggregation of its local model with the received models. Roy \textit{et al.} \cite{roy2019braintorrent} introduced the BrainTorrent framework, in which a requesting client communicates with all clients to obtain information about available model versions, and clients with new versions send their models to the requesting client for aggregation.

\subsection{Variant: Star}
\label{Sec. Variant: Star}

The star variant resembles the CFL model, where one client assumes the role of the server to coordinate and interact with other clients, as depicted in Fig. \ref{Fig. Network topology}(d) and Fig. \ref{Fig. World Network}(c). The star variant operates in two different modes. In the first mode, similar to CFL, the central client is responsible for receiving, aggregating, and distributing the local models. However, unlike CFL, the central client also generates original data and utilizes the models for perception and decision-making. This mode emphasizes a family-like relationship, where one member has more computational and communication power to assist the other clients. In the second mode of operation, the focus is on geographic interoperability among clients. As some clients in the community are geographically dispersed, there is a client that serves as the geographical center for these clients. To conserve communication resources, the surrounding clients transmit their models to the central client, which then forwards the models to the other clients.

Pappas \textit{et al.} \cite{pappas2021ipls} introduced a split learning framework within a star DFL architecture, where clients train different layers of a model and update the model parameters with the central client. This approach allows for distributed model training and collaboration among clients. Another example of a star variant is the Swarm Learning framework proposed by Warnat-Herresthal \textit{et al.} \cite{warnat2021swarm}, which involves the dynamic election of a leader to aggregate model parameters. In Swarm Learning, the leader plays a central role in coordinating the aggregation process and facilitating collaboration among the clients.

\subsection{Variant: Hybrid}
\label{Sec. Variant: Hybrid}
The Hybrid variant of DFL encompasses a wide range of configurations, combining elements from various other variants. It is considered the most promising option for practical applications due to its adaptability to different scenarios \cite{wang2021network,wang2021device,wang2023device}. However, the complexity of configuring a hybrid variant can pose challenges. One example of a hybrid variant, as depicted in Fig. \ref{Fig. Network topology}(g), involves connecting two ring variants through a central client. In this configuration, the hybrid variant provides global connectivity, allowing for the sharing of client models and knowledge within the framework. The two ring variants can also be treated as a single entity, with only one communication channel connected to the two central clients. Another illustration of a hybrid variant, shown in Fig. \ref{Fig. World Network}(d), involves dividing clients into organizations based on geographical locations (e.g., continents). Within each organization, a mesh topology network is established, and a leader is elected. These leaders then form a ring topology network among themselves. The hybrid variants do not have a fixed structure and can be customized to meet the specific requirements of real-world scenarios. The hybrid variant offers several advantages.

Firstly, the hybrid variant helps in saving communication resources. This is achieved through the knowledge dissemination between the leaders of two organizations, where only the aggregated global model is shared. By transmitting only the essential information, the hybrid variant reduces the communication overhead. Additionally, organized knowledge dissemination further enhances resource efficiency by minimizing the sharing of irrelevant or invalid information. This approach is particularly advantageous when establishing communication between two geographically distant organizations, as the single-line connection reduces the resource requirements for long-distance communication. Considering that the communication between organizations represents the dissemination of knowledge across states, countries, and continents \cite{gerla1977topological}, the clients representing the institutions establish a stable and well-structured communication connection to facilitate the exchange of knowledge between their respective organizations.

Secondly, the hybrid variant offers enhanced security. With two central clients in control, they have the ability to unilaterally disconnect the communication between organizations, ensuring the protection of their respective knowledge from potential leaks or unauthorized access. This adds an extra layer of security to the DFL system.

Thirdly, the hybrid variant provides a more personalized approach. Each organization's aggregated model is organization-specific, tailored to the unique characteristics of its local data. This personalized model may offer better applicability to the specific needs and requirements of the organization. While model knowledge is shared between the two organizations, the decision of whether to utilize the other organization's model is subject to further investigation and discussion. By thoroughly assessing the performance of the other organization's model, clients can ensure that their own model remains uncontaminated and unaffected by potentially inferior or incompatible models.

Xing \textit{et al.} \cite{xing2020decentralized} proposed a hybrid DFL network that establishes connections only with neighboring clients, and model parameters are broadcast-gossiped only among these neighboring clients. Their approach takes into account various factors such as link blockages, channel fading, and mutual interference, to ensure efficient and reliable communication. Building upon this work, Shi \textit{et al.} \cite{shi2021over} further improved the convergence performance by incorporating coding strategies, gradient tracking, and variance reduction algorithms. In a similar vein, Wang \textit{et al.} \cite{wang2022matcha} developed a dynamic hybrid DFL framework called Matcha. Matcha introduces the concept of creating different network topologies at each iteration to enhance convergence speed. The algorithm consists of two main parts. Firstly, an initial network topology pre-processing step where Matcha performs matching decomposition on a base communication topology to obtain disjoint sub-graphs, including sub-graphs with only two-peer connections. Next, matching activation probabilities are computed to maximize the connectivity of the graph, and a new random topology graph is generated for each iteration. The key idea behind Matcha is to achieve faster convergence by enabling more frequent communication on connectivity-critical links (e.g., central clients) and reducing communication latency by decreasing the frequency of communication on other connections. Matcha is particularly advantageous for hybrid networks with unknown or dynamic central clients. However, it may not exhibit the same advantages in scenarios involving research institutions where central clients are known and pre-determined.

\section{Challenge and Potential Solutions in DFL}
\label{Sec. Challenge in DFL}
Based on the current SOTA technology, this section aims to discuss and analyze potential challenges and future research directions for DFL. Additionally, the variants mentioned in Section \ref{Sec. Variants of DFL} can be regarded as potential solutions to address these challenges.

\subsection{High Communication Overhead}
\label{Sec. High Communication Overhead}
DFL is widely recognized as an extremely communication resource-efficient approach compared to CFL. However, researchers are still striving for further savings in communication resources and reduced communication complexity \cite{kalra2023decentralized}. We discuss the existing CFL frameworks in Section \ref{Sec. Variant of CFL}, and we consider introducing viable methods to achieve efficient communication in DFL. Wang \textit{et al.} \cite{wang2022accelerating} introduced a method called optimization of topology construction and model compression (CoCo) that aims to improve communication efficiency and convergence speed in DFL. CoCo achieves this by employing adaptive techniques for constructing the DFL network topology and assigning an appropriate model compression ratio to each participating client. It achieves this by adaptively constructing the DFL network topology and assigning an appropriate model compression ratio to each client. 

In addition to model compression, it is also important to investigate how to leverage efficient communication lines and reduce the overall communication length. Variants such as star and hybrid variants, which select geocentric clients and resource-rich clients as leaders, have been proven to be effective solutions in this regard. Some researchers have also focused on addressing the bandwidth differences among different communication lines \cite{tang2020communication,zhou2021communication}. It is worth noting that the dynamic hybrid variant proposed by Wang \textit{et al.} \cite{wang2022matcha} emphasizes the importance of communication efficiency and suggests frequent communication with key clients to achieve faster convergence. A considerable body of research emphasizes the importance of efficient communication in DFL and proposes various strategies and methods to reduce complexity and optimize communication resources. Further exploration in this direction is expected to facilitate the potential deployment of DFL frameworks in real-world applications.

\subsection{Computational and Storage Burden}
\label{Sec. Computational and Storage Burden}
Compared to the CFL and \texttt{Continual} paradigm, the \texttt{Aggregate} paradigm imposes significantly higher demands on client-side computational and storage resources. As there is no dedicated server in the \texttt{Aggregate} paradigm, clients are responsible for storing previous model parameters and performing aggregation computations alongside local model training. Consequently, the computational and storage burdens pose challenges for client hardware.

One potential solution is to adopt the transfer learning concept and fix the weights of the lower layers in all models. In this approach, the lower layers serve as feature extractors for a specific task and are expected to be similar across models, while the higher-level representations remain task-specific. By fixing these parameters, there is no need for gradient descent, aggregation computations, or communication-related to these layers. Moreover, this approach reduces storage requirements, thereby substantially mitigating the resource consumption of the client. Currently, with the widespread availability of high-performance GPU computing resources, the challenges related to computational and storage burdens are gradually diminishing. This is especially true in DFL scenarios where institutions and organizations serve as clients. However, in contexts such as mobile services dominated by smartphones and on-board units in vehicular edge devices, there is still value in researching ways to reduce computational complexity and optimize storage efficiency.

\subsection{Vulnerability in Cybersecurity}
\label{Sec. Vulnerability in Cybersecurity}
Network security has always been a major challenge in FL, and this challenge is particularly prominent in DFL \cite{rahman2020internet,alazab2021federated,ghimire2022recent,zhou2023decentralized}. In the traditional CFL setting, clients communicate with a central server, typically operated by a research institution or a large commercial organization. While there is still potential for attacks and data poisoning between clients and the server, communication is generally more regulated and protected compared to DFL. In DFL, the knowledge exchange occurs directly among users within a local area network, with free and unrestricted sharing agreements, which poses an increased risk of privacy exposure. Malicious attacks from clients, poisoned data, free-riding attacks, and other malicious behaviors are all possible in this decentralized setting \cite{kuo2022detecting,wang2023enhancing}.

Kuo \textit{et al.} \cite{kuo2018modelchain} proposed the integration of blockchain into a decentralized learning framework to enhance privacy protection, which can also be applied to DFL. Chen \textit{et al.} \cite{chen2018machine} integrated a differential privacy mechanism based on blockchain technology. Bellet \textit{et al.} \cite{bellet2018personalized} introduced an asynchronous and differential privacy algorithm in DFL to safeguard user privacy. He \textit{et al.} \cite{he2019central} addressed trust issues between clients by employing an online push-sum algorithm to actively push local models to trusted clients. Shayan \textit{et al.} \cite{shayan2020biscotti} proposed the Biscotti DFL system, which incorporates multiple privacy and security protection techniques, including the Multi-Krum defense to prevent poisoning attacks, differential privacy noise to protect privacy, and secure aggregation. The future research direction in cybersecurity will involve the roles of attackers and defenders, focusing on developing targeted attack and defense mechanisms for different DFL variants.

\subsection{Lack of Incentive Mechanism}
In the absence of server management, the issue of fairness in aggregation has been effectively addressed in DFL. However, the lack of incentives and mutual distrust among clients can significantly impact their willingness to contribute knowledge. A key issue is the lack of incentives, which may lead to free-riding attacks where clients choose to benefit from the models without contributing their own knowledge \cite{witt2022decentral,yuan2023digital}.

In the context of DFL, the feasibility of incentive mechanisms based on game theory, such as Stackelberg games \cite{khan2020federated}, raises questions due to the requirements on game leaders, participants, and rewards. One potential solution could be the integration of reputation-based incentive mechanisms using blockchain and smart contracts. Kang \textit{et al.} \cite{kang2019incentive} proposed assigning reputation scores to clients to represent and quantify their reliability. Clients with higher contributions and reputations can receive greater rewards. However, designing effective and practical incentive mechanisms for DFL remains an open problem.

In cases where task providers do not exist or there are no explicit rewards, punitive incentives may also be a potential solution. Clients who fail to contribute or engage in malicious behavior could face penalties or reduced access to the benefits of the DFL framework. Further research is needed to explore and develop robust incentive mechanisms tailored specifically for DFL systems. Designing effective incentive mechanisms to encourage active participation, foster trust, and stimulate enthusiastic knowledge sharing will greatly facilitate the dissemination of knowledge in DFL.

\subsection{Lack of Management}
\label{Sec. Lack of Management}

In DFL, the absence of a central server for managing all clients poses a significant challenge in receiving and sharing knowledge in an organized manner. The lack of central management can lead to confusion, particularly among clients with varying sample sizes, computational resources, and communication capabilities. In the ring variant, a client only needs to wait for the model parameters from the previous client, while in the mesh variant, a client needs to wait for model parameters from all other clients. Such dependencies on other clients for model transmission can result in deadlocks, causing the entire system to halt. Moreover, the communication among clients may not be robust, considering the possibility of SPoF. The absence of management is particularly problematic in the hybrid variant depicted in Fig. \ref{Fig. World Network}(d), where clients communicate globally. The lack of management can lead to reduced operational efficiency, confusion regarding model versions, and performance degradation.

To address the challenge of lack of management in DFL, researchers have proposed several approaches. One approach is to pre-request the status of other clients, such as their model versions, before initiating knowledge transfer \cite{roy2019braintorrent}. By obtaining accurate information, a client can then request the transfer of the entire model data. Some star variants enforce the knowledge dissemination flow among clients by designating a leader \cite{yu2020proactive}. This leader is responsible for regulating knowledge dissemination among the remaining clients. Additionally, Chen \textit{et al.} \cite{chen2021bdfl} introduced the BDFL framework, a mesh DFL framework specifically designed for autonomous vehicles. In this framework, a leader is randomly selected in each communication round, offering advantages such as increased privacy and security protection against Byzantine faults, as well as enhanced management through the leader's command issuance. In real-world scenarios, clients may face challenges where they lack knowledge about each other's statuses, leading to issues such as model version discrepancies and even system paralysis, such as in the case of an SPoF. Therefore, future research directions aim to ensure the smooth operation of the system by incorporating additional information or establishing contingency plans. These measures can help mitigate the impact of uncertainty and improve the reliability and robustness of the DFL framework.

\section{Conclusion}
\label{Sec. Conclusion}
In this paper, we provided an extensive exploration of the DFL framework, covering communication protocols, network topologies, paradigm proposals, extension variants, challenges, and potential solutions. Our aim is to offer a comprehensive, well-defined, and systematic perspective that organizes and synthesizes the existing literature and definitions, thereby facilitating a comprehensive introduction to DFL for new researchers. Given that DFL is a rapidly evolving area, we established a solid theoretical foundation by defining and discussing five variants in this paper. This not only provides researchers with a comprehensive understanding of the field but also fosters the generation of new ideas and collaborations among peers. 

It is important to note that our approach differs from traditional surveys, as we presented our own insights and innovative thinking on DFL. Moreover, this paper uncovers a considerable number of previously unexplored types within the DFL framework. For example, no existing studies have demonstrated the integration of the \texttt{Continual} paradigm with mesh network topology. Researchers might consider asynchronous DFL, where clients acquire data and learn at different times, thus benefiting from the dual advantages offered by the \texttt{Continual} paradigm and the mesh network topology, such as reduced communication overhead, personalization, and more. By considering diverse usage scenarios, we aim to stimulate and extend the research interest of other DFL practitioners, enabling them to adapt the framework to their specific needs.

\bibliographystyle{IEEEtran}
\small\bibliography{reference}

\vfill

\begin{IEEEbiography}
[{\includegraphics[width=1in,height=1.25in,clip,keepaspectratio]{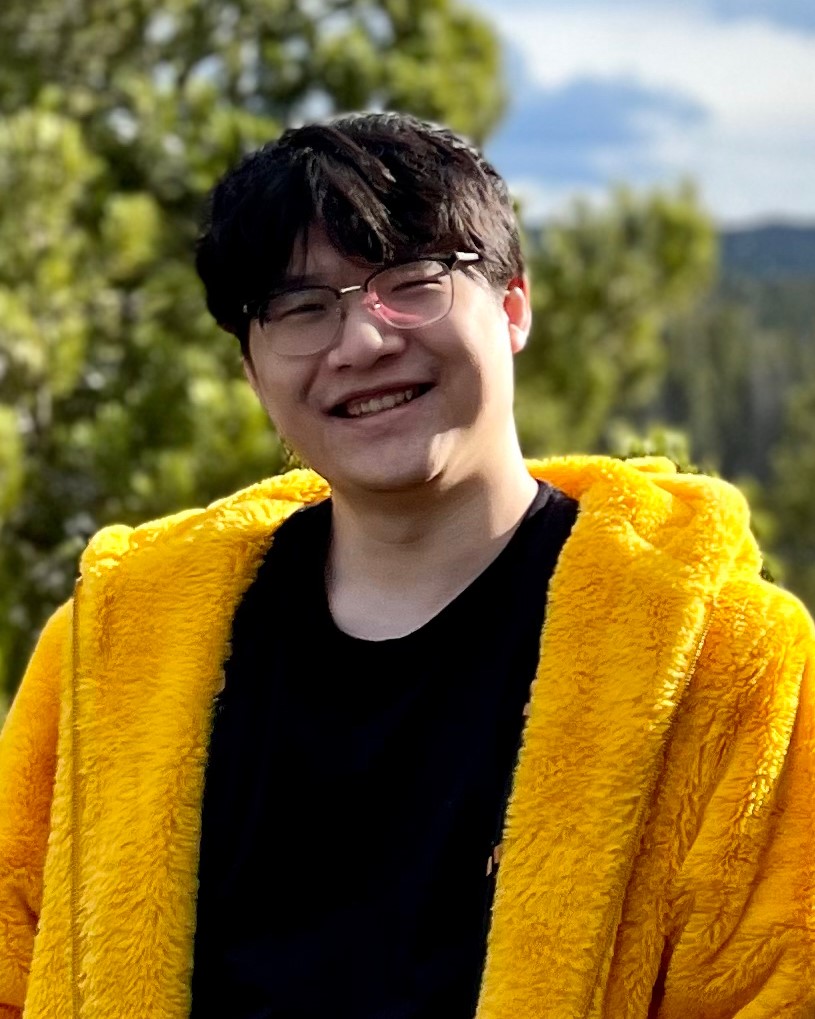}}]
{Liangqi Yuan}
(S'22) received the B.E. degree from the Beijing Information Science and Technology University, Beijing, China, in 2020, and the M.Sc. degree from the Oakland University, Rochester, MI, USA, in 2022. He is currently pursuing the Ph.D. degree with the School of Electrical and Computer Engineering, Purdue University, West Lafayette, IN, USA. His research interests are in the areas of sensors, the internet of things, human–computer interaction, signal processing, and machine learning.
\end{IEEEbiography}

\begin{IEEEbiography}
[{\includegraphics[width=1in,height=1.25in,clip,keepaspectratio]{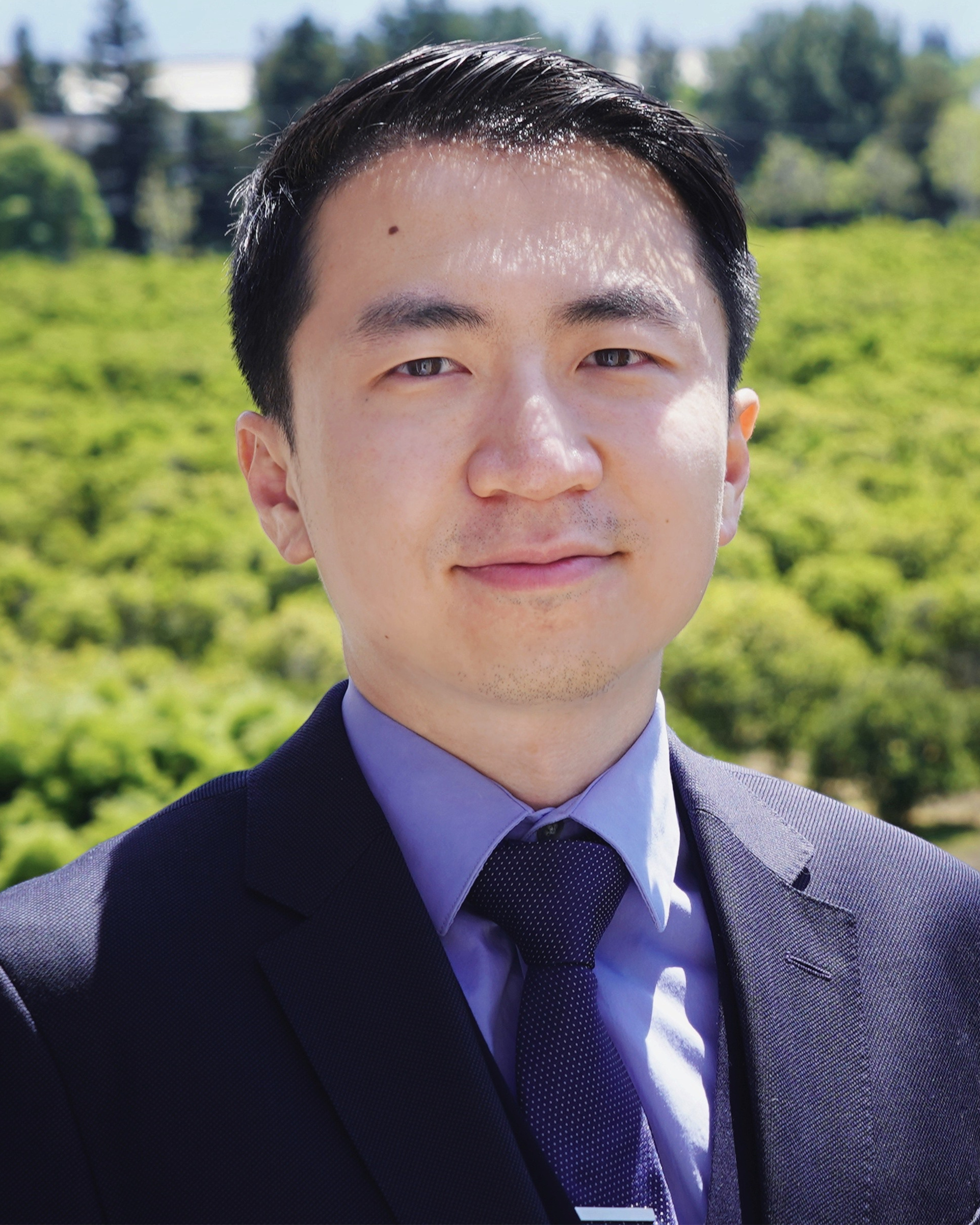}}]
{Ziran Wang}
(S'16-M'19) received his Ph.D. degree in Mechanical Engineering from the University of California, Riverside in 2019. He is a tenure-track assistant professor in the College of Engineering at Purdue University, where he leads the Purdue Digital Twin Lab. Prior to this, Dr. Wang was a principal researcher at Toyota Motor North America in Mountain View, California. 

Dr. Wang serves as the founding chair of the IEEE Technical Committee on Internet of Things in Intelligent Transportation Systems, a member of three other IEEE technical committees, and a technical program committee member of multiple IEEE and ACM conferences. Dr. Wang is an associate of four academic journals, including IEEE Transactions on Intelligent Vehicles and IEEE Internet of Things Journal. His research achievements have been demonstrated at the Consumer Electronics Show (CES), and acknowledged by the U.S. Department of Transportation Dissertation Award, the IEEE ``Shape the Future of ITS'' 1st Prize Award, and five other best paper awards from IEEE and SAE. He is an author of five book chapters, 50+ refereed papers, and 50+ patent applications. His research focuses on autonomous driving, human-autonomy teaming, and digital twin. 
\end{IEEEbiography}

\begin{IEEEbiography}
[{\includegraphics[width=1in,height=1.25in,clip,keepaspectratio]{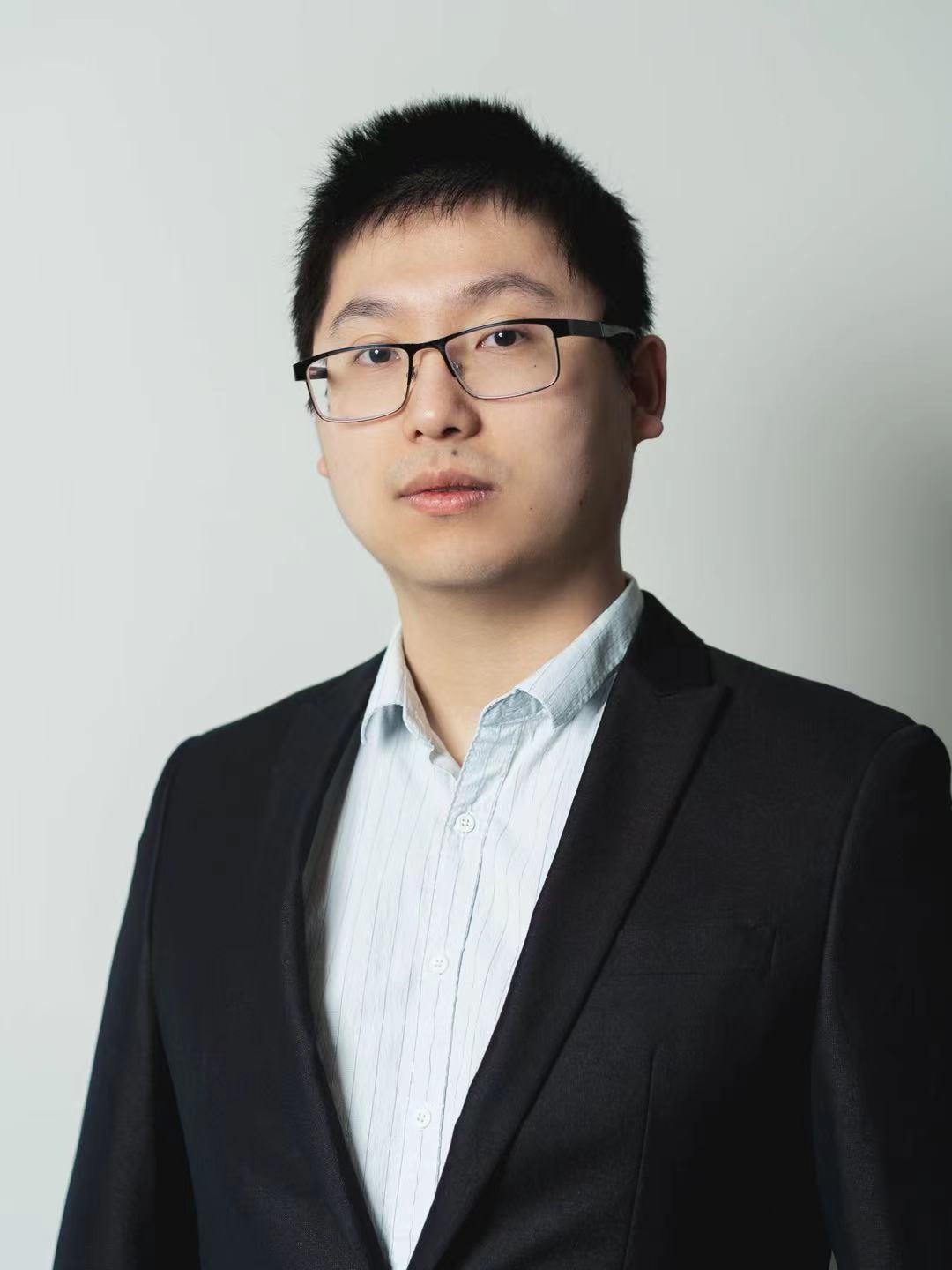}}]
{Lichao Sun} received the Ph.D. degree in computer science from the University of Illinois Chicago, Chicago, IL, USA, in 2020, under the supervision of Prof. Philip S. Yu.

He is currently an Assistant Professor with the Department of Computer Science and Engineering, Lehigh University, Bethlehem, PA, USA. He has published more than 45 research articles in top conferences and journals, such as CCS, USENIXSecurity, NeurIPS, KDD, ICLR, the Advancement of AI (AAAI), the International Joint Conference on AI (IJCAI), ACL, NAACL, TII, TNNLS, and TMC. His research interests include security and privacy in deep learning and data mining. He mainly focuses on artificial intelligence (AI) security and privacy, social networks, and NLP applications.
\end{IEEEbiography}

\begin{IEEEbiography}
[{\includegraphics[width=1in,height=1.25in,clip,keepaspectratio]{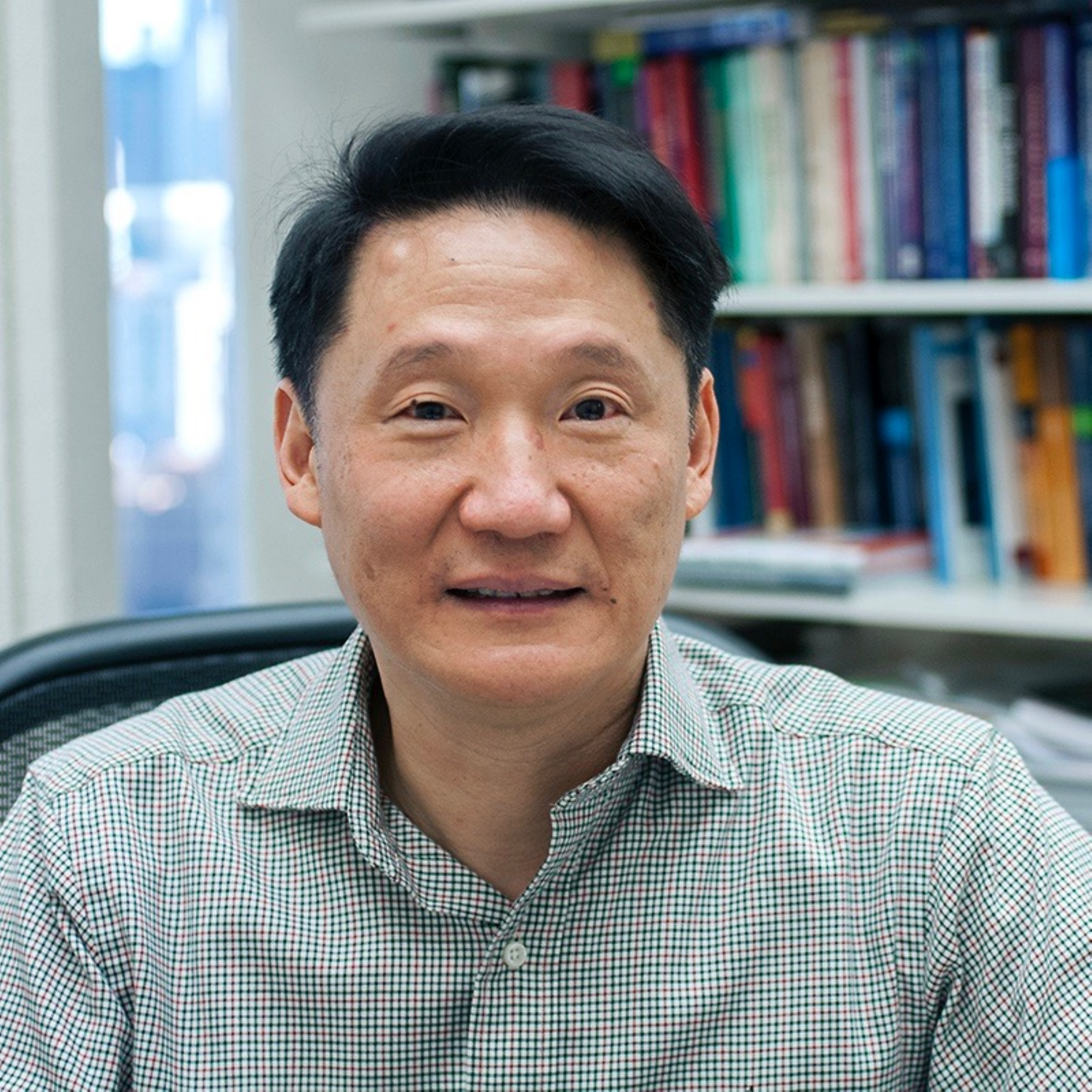}}]
{Philip S. Yu} (Life Fellow, IEEE) received the B.S. degree in electrical engineering (E.E.) from the National Taiwan University, New Taipei, Taiwan, in 1992, the M.S. and Ph.D. degrees in E.E. from Stanford University, Stanford, CA, USA, in 1976 and 1978, respectively, and the M.B.A. degree from New York University, New York, NY, USA, in 1982.

He is currently a Distinguished Professor of computer science with the University of Illinois Chicago (UIC), Chicago, IL, USA, and also holds the Wexler Chair in Information Technology. Before joining UIC, he was with IBM, USA, where he was the Manager of the Software Tools and Techniques Department, Watson Research Center. He has published more than 1200 papers in refereed journals and conferences. He holds or has applied for more than 300 U.S. patents. His research interest is on big data, including data mining, data stream, database, and privacy.

Dr. Yu is a fellow of the ACM. He was a recipient of the ACM SIGKDD 2016 Innovation Award for his influential research and scientific contributions to mining, fusion, and anonymization of big data, the IEEE Computer Society’s 2013 Technical Achievement Award for pioneering and fundamentally innovative contributions to the scalable indexing, querying, searching, mining, and anonymization of big data, and the Research Contributions Award from IEEE International Conference on Data Mining (ICDM) in 2003 for his pioneering contributions to the field of data mining. He received the ICDM 2013 10-Year Highest-Impact Paper Award and the EDBT Test of Time Award in 2014. He was the Editor-in-Chief of ACM Transactions on Knowledge Discovery from Data from 2011 to 2017 and IEEE Transactions on Knowledge and Data Engineering from 2001 to 2004.
\end{IEEEbiography}

\begin{IEEEbiography}
[{\includegraphics[width=1in,height=1.25in,clip,keepaspectratio]{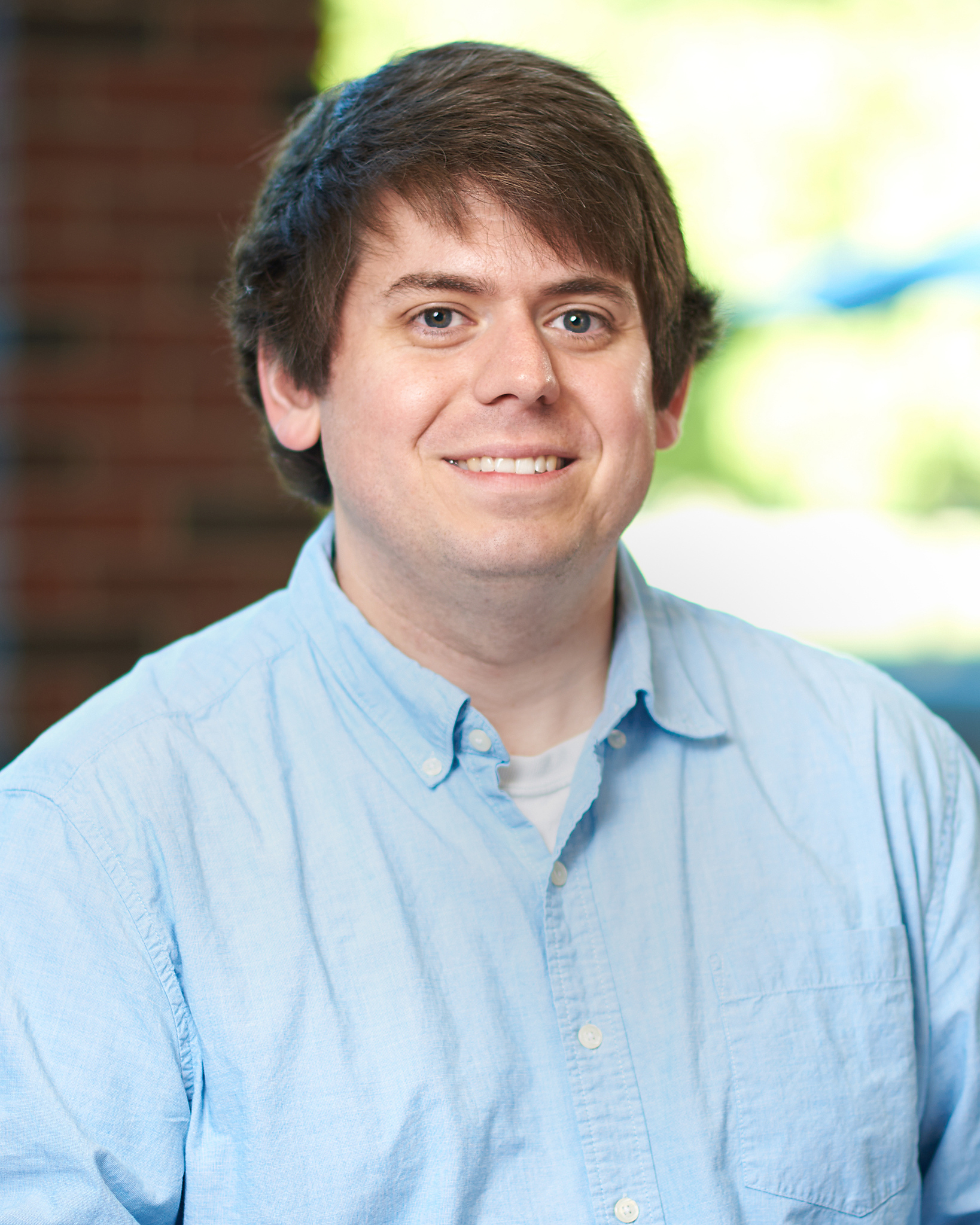}}]
{Christopher G. Brinton}
(S'08, M'16, SM'20) is the Elmore Rising Star Assistant Professor of Electrical and Computer Engineering (ECE) at Purdue University. His research interest is at the intersection of networking, communications, and machine learning, specifically in fog/edge network intelligence, distributed machine learning, and data-driven wireless network optimization. Dr. Brinton is a recipient of the NSF CAREER Award, ONR Young Investigator Program (YIP) Award, DARPA Young Faculty Award (YFA), Intel Rising Star Faculty Award, and roughly \$15M in sponsored research projects as a PI or co-PI. He has also been awarded Purdue College of Engineering Faculty Excellence Awards in Early Career Research, Early Career Teaching, and Online Learning. He currently serves as an Associate Editor for IEEE/ACM Transactions on Networking. Prior to joining Purdue, Dr. Brinton was the Associate Director of the EDGE Lab and a Lecturer of Electrical Engineering at Princeton University. He also co-founded Zoomi Inc., a big data startup company that has provided learning optimization to more than one million users worldwide and holds US Patents in machine learning for education. His book The Power of Networks: 6 Principles That Connect our Lives and associated Massive Open Online Courses (MOOCs) have reached over 400,000 students to date. Dr. Brinton received the PhD (with honors) and MS Degrees from Princeton in 2016 and 2013, respectively, both in Electrical Engineering.
\end{IEEEbiography}

\end{document}